%% file: main.tex
\RequirePackage{fix-cm}
\documentclass[twocolumn]{svjour3}    % twocolumn
\pdfoutput=1

\usepackage{graphicx}
\usepackage{natbib}
\usepackage{marvosym}
\usepackage{url}            % simple URL typesetting
\usepackage{booktabs}       % professional-quality tables
\usepackage{amsfonts}       % blackboard math symbols
\usepackage{nicefrac}       % compact symbols for 1/2, etc.
\usepackage{microtype}      % microtypography
\usepackage{graphicx}
\usepackage{enumitem}
\usepackage{graphicx}    % For \resizebox
\usepackage{colortbl}    % For \rowcolor
\usepackage{xcolor}      % For color definitions (gray!12)
\usepackage[table]{xcolor} % 支持表格行颜色（\rowcolor）
\usepackage{graphicx}      % 支持表格缩放（\resizebox）
\usepackage{bm}
\usepackage{nccmath}
\usepackage{boldline}
\usepackage{booktabs}        % To thicken table lines
\usepackage{threeparttable}

\usepackage[noend]{algcompatible}

\usepackage{capt-of}
\usepackage{color,soul}
\usepackage[pagebackref=false,breaklinks=true,colorlinks,bookmarks=false,urlcolor=magenta, citecolor=blue]{hyperref}

\usepackage{times}
\usepackage{epsfig}
\usepackage{multirow}
\usepackage{amsmath}
\usepackage{amssymb}

\usepackage{bbding}
\usepackage{wrapfig}
\usepackage{comment}
\usepackage{pifont}
\usepackage[table]{xcolor}
\usepackage{bm}
\usepackage{soul}

\usepackage{stfloats}
\usepackage{cuted}
\usepackage{capt-of}
\usepackage{graphicx}
\usepackage{booktabs}
\usepackage{enumitem}
\usepackage[table]{xcolor} % 支持表格行颜色（\rowcolor）

\usepackage[normalem]{ulem}
\usepackage{graphicx}      % 支持表格缩放（\resizebox）
\usepackage{amsfonts}
\usepackage{bbm}
\usepackage{textcomp}

\usepackage[linesnumbered,ruled,vlined]{algorithm2e}
\usepackage{makecell}
\usepackage{subfigure}
\usepackage{arydshln}

\usepackage{xspace}
\usepackage{paralist}
\input{preamble}

\def\cm#1{\checkmark}

\newcommand{\etal}{\textit{et al}. }
\newcommand{\ie}{\textit{i}.\textit{e}., }
\newcommand{\eg}{\textit{e}.\textit{g}., }

\useunder{\underline}{\ul}{}

\usepackage{tikz}

\begin{document}
\begin{sloppypar}
\title{OmniTrack++: Omnidirectional Multi-Object Tracking by Learning Large-FoV Trajectory Feedback}

\author{
Kai Luo$^{1}$ \and 
Hao Shi$^{2,3}$ \and 
Kunyu Peng$^{4,5}$ \and 
Fei Teng$^{1}$ \and 
Sheng Wu$^{1}$ \and 
Kaiwei Wang$^{2}$ \and   
Kailun Yang$^{1,\dagger}$
}

\institute{
Kailun Yang (kailun.yang@hnu.edu.cn) \\ 
$^1$. {\orgname{Hunan University}, \city{Changsha}, \country{China}.} \\
$^2$. {\orgname{Zhejiang University}, \city{Hangzhou}, \country{China}.}  \\
$^3$. {\orgname{Ant Group}, \city{Hangzhou}, \country{China}.}  \\
$^4$. {\orgname{Karlsruhe Institute of Technology}, \city{Karlsruhe}, \country{Germany}.}  \\
$^5$. {\orgname{INSAIT, Sofia University ``St. Kliment Ohridski''}, \city{Sofia}, \country{Bulgaria}.}  \\
$\dagger$ Corresponding author.\\
}

\date{Received: date / Accepted: date}

\maketitle

\begin{abstract}
To address panoramic distortion, large search space, and identity ambiguity under a $360^\circ$ FoV, OmniTrack++ adopts a feedback-driven framework that progressively refines perception with trajectory cues. A DynamicSSM block first stabilizes panoramic features, implicitly alleviating geometric distortion. On top of normalized representations, FlexiTrack Instances use trajectory-informed feedback for flexible localization and reliable short-term association. To ensure long-term robustness, an ExpertTrack Memory consolidates appearance cues via a Mixture-of-Experts design, enabling recovery from fragmented tracks and reducing identity drift. Finally, a Tracklet Management module adaptively switches between end-to-end and tracking-by-detection modes according to scene dynamics, offering a balanced and scalable solution for panoramic MOT. To support rigorous evaluation, we establish the EmboTrack benchmark, a comprehensive dataset for panoramic MOT that includes QuadTrack, captured with a quadruped robot, and BipTrack, collected with a bipedal wheel-legged robot. Together, these datasets span wide-angle environments and diverse motion patterns, providing a challenging testbed for real-world panoramic perception. 
Extensive experiments on JRDB and EmboTrack demonstrate that OmniTrack++ achieves state-of-the-art performance, yielding substantial HOTA improvements of $+3.94 (21.56 \rightarrow 25.50)$ on JRDB and $+15.03 (19.87\rightarrow34.90)$ on QuadTrack over the original OmniTrack. These results highlight the effectiveness of trajectory-informed feedback, adaptive paradigm switching, and robust long-term memory in advancing panoramic multi-object tracking. Datasets and code will be made available at \url{https://github.com/xifen523/OmniTrack}.

\end{abstract}
%%%%%%%%%%%%%%%%%%%%%%%%%%%%%%%%%%%%%%%%%%%%%%%%%%%%%%%%%%%%%%%%%%%%%%%%%

\section{Introduction}\label{sec1}

\input{Sec_Introduction}

% \vspace{-15px}
\section{Related Work}\label{sec2}
\input{Sec_Related_Work}

\section{OmniTrack++: Proposed Framework}\label{sec3}
\input{Sec_Method}

\section{EmboTrack: a Dynamic 360{\textdegree} MOT Dataset}\label{sec4}

\input{Sec_EmboTrack}

\section{Experiment Results and Analysis}\label{sec5}
\input{Sec_Experiments}

\section{Conclusion}\label{sec6}
\input{Sec_Conclusion}

\vspace{-4px}
\section*{Conflict of Interest}
\vspace{-2px}

The authors declare no potential conflicts of interest with respect to the research, authorship, and publication of this article. All authors have participated in the research and have reviewed the final version of the manuscript. 

% \vspace{-4 px}
\section*{Acknowledgments}
% \vspace{-3 px}
This work was supported in part by the National Natural Science Foundation of China (Grant No. 62473139), in part by the Hunan Provincial Research and Development Project (Grant No. 2025QK3019), and in part by the State Key Laboratory of Autonomous Intelligent Unmanned Systems (the opening project number ZZKF2025-2-10).

\bibliographystyle{sn-basic}      %
\bibliography{sn-bibliography}

\clearpage

\end{sloppypar}  
\end{document}

%% file: preamble.tex
\newcommand{\city}[1]{#1}
\newcommand{\country}[1]{#1}
\newcommand{\orgname }[1]{#1}

\newcommand{\name}[0]{OmniTrack}
\usepackage{makecell}
\usepackage[compatibility=false]{caption}
\makeatletter
\newcommand{\algorithmfootnote}[2][\footnotesize]{%
  \let\old@algocf@finish\@algocf@finish% Store algorithm finish macro
  \def\@algocf@finish{\old@algocf@finish% Update finish macro to insert "footnote"
    \leavevmode\rlap{\begin{minipage}{\linewidth}
    #1#2
    \end{minipage}}%
  }%
}
\makeatother

\makeatletter
\newcommand{\shortcline}[1]{%
  \noalign{\global\advance\@totalleftmargin by 0.15em}% 调整左缩进
  \noalign{\global\advance\linewidth by -10em}% 调整右缩进
  \cline{#1}%
  \noalign{\global\advance\@totalleftmargin by -0.5em}% 恢复左缩进
  \noalign{\global\advance\linewidth by 1em}% 恢复右缩进
}
\makeatother

\definecolor{codegreen}{rgb}{0.0,0.6,0.0}
\definecolor{mygray}{gray}{.9}
\definecolor{mygray1}{gray}{.7}
\definecolor{tabgray}{rgb}{0.957,0.945,0.925}
\definecolor{tabgray2}{rgb}{0.90,0.85,0.80}
\definecolor{poster_color1}{rgb}{1,1,0.949}
\definecolor{poster_color2}{rgb}{0.925,0.980,0.898}
\definecolor{poster_color3}{rgb}{0.865, 0.925, 0.835}

\newcommand{\car}{\includegraphics[width=4mm]{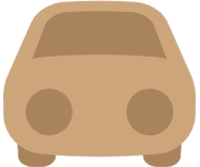}}
\newcommand{\robot}{\includegraphics[width=4mm]{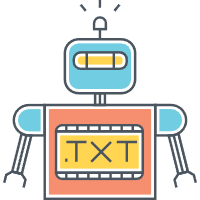}}
\newcommand{\robotdog}{\includegraphics[width=4mm]{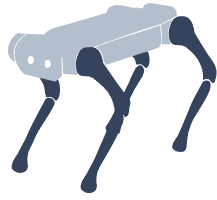}}
\newcommand{\webm}{\includegraphics[width=3.6mm]{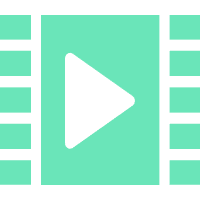}}
\newcommand{\mywebm}{\includegraphics[width=3.5mm]{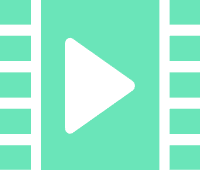}}

\newcommand{\wheels}{\includegraphics[width=4mm]{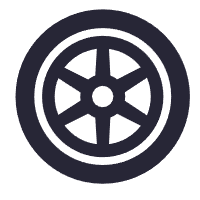}}
\newcommand{\gait}{\includegraphics[width=4mm]{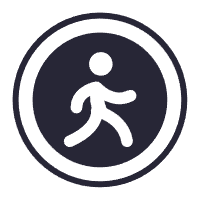}}
\newcommand{\stationary}{\includegraphics[width=4mm]{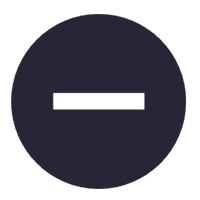}}

\newcommand{\mycheckmark}{\includegraphics[width=3.5mm]{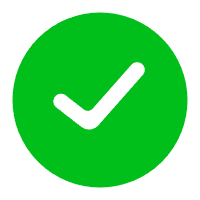}}
\newcommand{\crossmark}{\includegraphics[width=3.5mm]{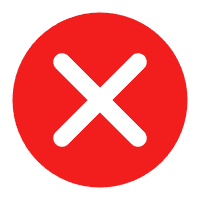}}
\newcommand{\wheelleg}{\includegraphics[width=3.5mm]{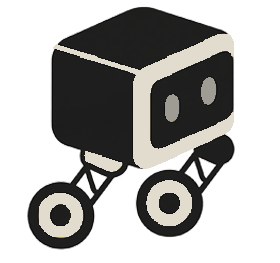}}

\newcommand{\topline}{\noalign{\hrule height 0.8 pt}} 
 
\newcommand{\bottomline}{\noalign{\hrule height 0.8 pt}} 

\definecolor{codebackgroundcolor}{RGB}{248,242,222}

\definecolor{r1}{rgb}{0.7569,0.2275,0.1294}
\definecolor{r2}{rgb}{0.0,0.5451,0.5451}
\definecolor{r3}{rgb}{0.5451,0.0,0.5451}

\definecolor{rblue}{rgb}{0,0.5,1}
\definecolor{hollywoodcerise}{rgb}{0.96, 0.0, 0.63}
\definecolor{lasallegreen}{rgb}{0.03, 0.47, 0.19}
\definecolor{hanpurple}{rgb}{0.32, 0.09, 0.98}
\definecolor{green(pigment)}{rgb}{0.0, 0.65, 0.31}

\definecolor{mygreen}{RGB}{20, 180, 70}
\definecolor{myorange}{RGB}{255, 140, 0}
\definecolor{myred}{RGB}{220, 30, 30}
\definecolor{mygray}{RGB}{211, 211, 211}

\hypersetup{
    breaklinks=true,
    colorlinks=true, 
    linkcolor={red},
    citecolor={hanpurple}, % 请确保已定义 hanpurple 颜色
    urlcolor={magenta}
}

%% file: Sec_Introduction.tex
Panoramic cameras, featuring a 360{\textdegree} Field of View (FoV), enable comprehensive perception of the surrounding environment~\cite{ai2022deep,xu2021spherical_dnns,zheng2025one_flight}. 
This unique capability renders them highly valuable across a range of vision-centric applications, including autonomous driving~\cite{Wen_2024_CVPR,cao2024occlusion}, robotic navigation~\cite{van2024visual,shi2023panoflow}, and human-computer interaction~\cite{wu2024effect,han2022panoramic_activity}.
In particular, for small-scale mobile platforms such as quadrupedal robots and bipedal wheel-legged robots, panoramic imaging offers a compact yet effective means to achieve full-scene situational awareness without the need for multiple sensors, thus reducing payload~\cite{shi2026oneocc,wu2025quadreamer,zhang2025humanoidpano,zhao2026panoramic}.

\begin{figure}[!t]
  \centering
  \includegraphics[width=0.48\textwidth]{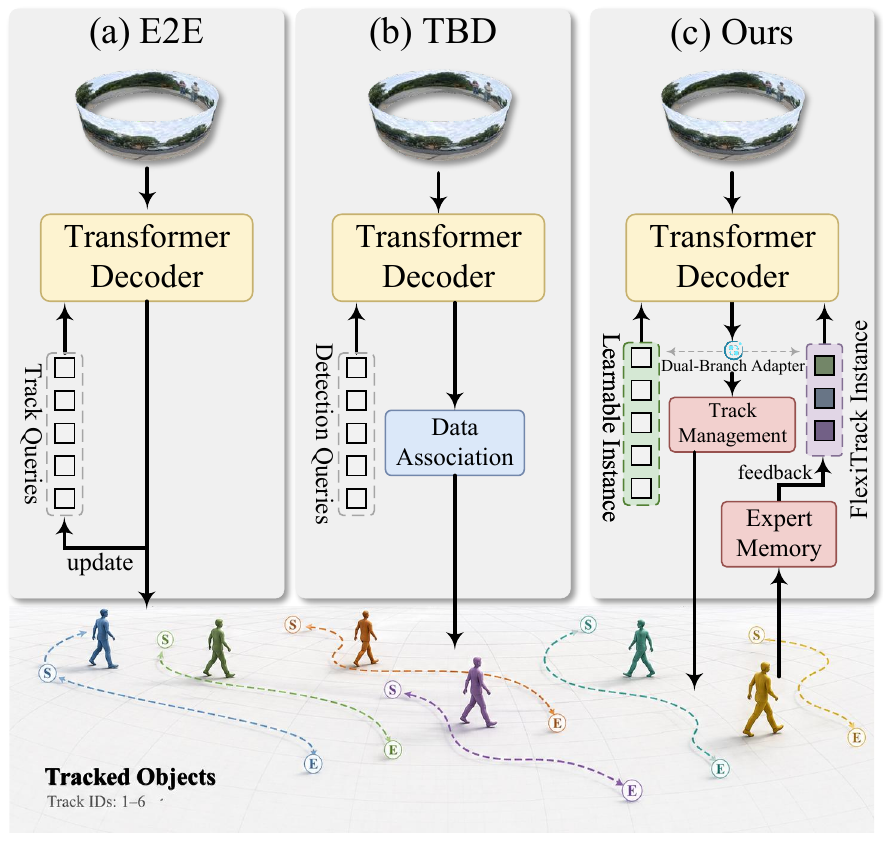}
  \caption{Comparison of mainstream tracking paradigms. (a) illustrates the typical End-To-End (E2E) paradigm; (b) shows the classical Tracking-By-Detection (TBD) paradigm; and (c) depicts our proposed \textbf{OmniTrack++} paradigm, which adaptively integrates and switches between the two paradigms. 
  In addition, OmniTrack++ employs a trajectory-feedback module that delivers rapid, large-FoV localization cues tailored to panoramic imagery, thereby narrowing the search space and stabilizing candidate selection, which ultimately improves data-association accuracy.}
  \label{fig:paradigm}
  % \vskip -2ex
\end{figure} 

Given their ability to capture holistic scene information, panoramic vision systems offer significant potential for enhancing perception in real-world environments~\cite{huang2024360loc,huang2023360vot}. 
However, effective panoramic perception requires not only comprehensive spatial coverage but also the ability to consistently interpret dynamic elements across time.
Among various perception techniques, Multi-Object Tracking (MOT)~\cite{bewley2016simple,wojke2017simple} stands out as a fundamental approach, as it jointly addresses object localization in space and association over time. MOT serves as a critical component in tasks such as scene understanding~\cite{li2024beyond,luo2024delving,xue2025usvtrack}, motion prediction~\cite{wang2025preformer,qin2023motiontrack}, and autonomous navigation planning~\cite{peng2024pnas,tosello2025temporal,hu2023_uniad} in complex, dynamic environments.

Despite the substantial progress in MOT, its application to panoramic imagery remains largely underexplored. 
Existing MOT algorithms~\cite{Chen_2024_CVPR,lv2024diffmot}, primarily designed for pinhole camera inputs, often fail to generalize well to panoramic settings due to intrinsic challenges, \eg, resolution degradation, geometric distortions, and non-uniform illumination when the images are unfolded into an equirectangular format~\cite{zheng2025one_flight}. 
These factors frequently lead to degraded performance—for instance, causing up to a $40\%$ increase in IDSWs~\cite{shen2024multi}—thus constraining the applicability of standard MOT pipelines to 360{\textdegree} panoramic scenarios.

\begin{figure*}[!t]
  \centering
  \includegraphics[width=0.98\textwidth]{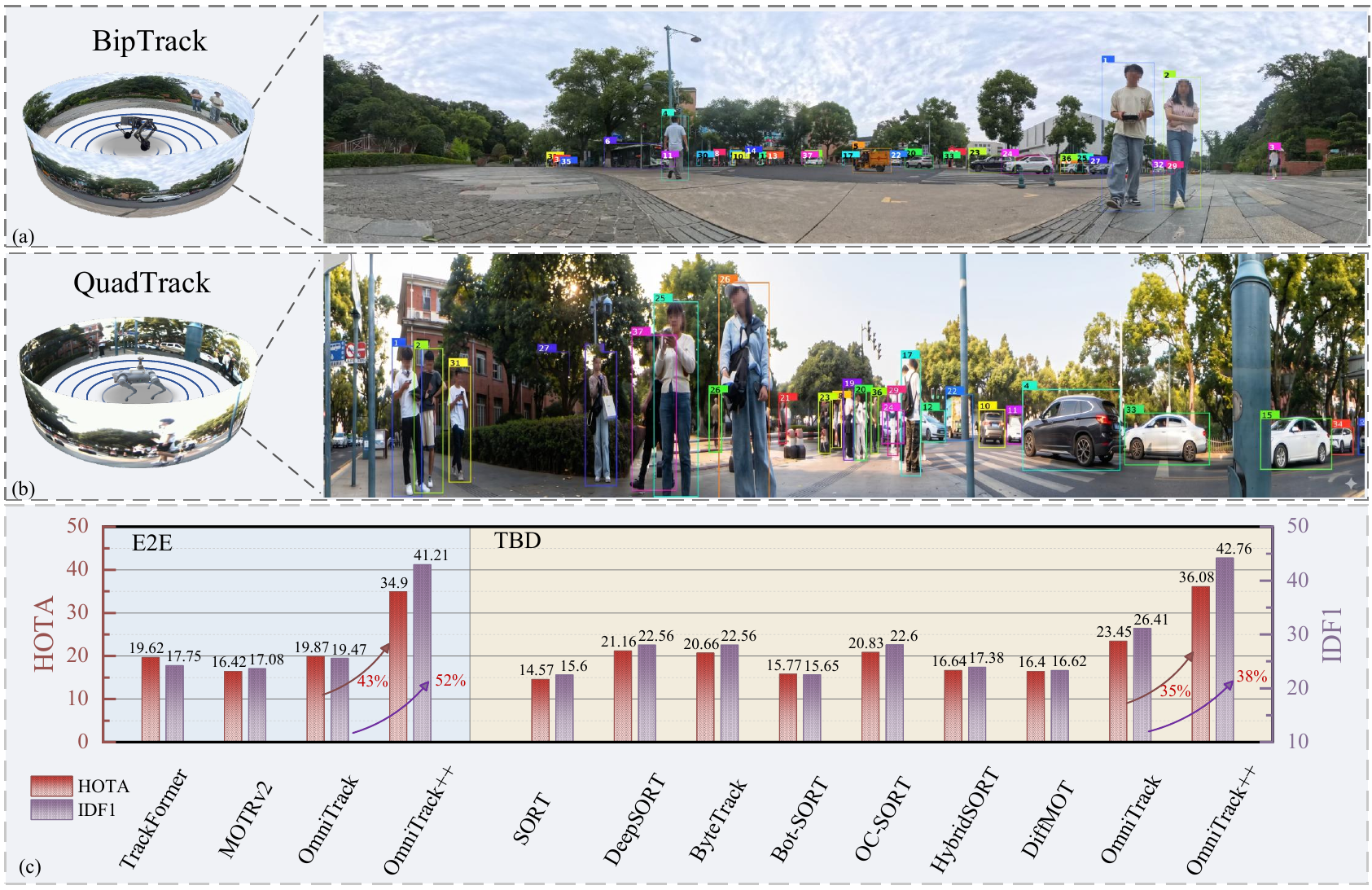}
  \caption{
  Overview of the EmboTrack benchmark (BipTrack and QuadTrack) and MOT results on the QuadTrack test set.
(a) BipTrack subset captured by a bipedal wheel–legged platform. (b) QuadTrack subset recorded by a quadrupedal platform. Dots under each object box indicate the large-FoV trajectory of the target, depicting its motion path within the panoramic scene. Both subsets provide panoramic MOT scenarios. (c) Quantitative comparison on QuadTrack: HOTA (left axis) and IDF1 (right axis) of representative MOT methods under E2E and TBD paradigms; OmniTrack++ achieves the highest overall accuracy.
  }
  \label{fig:overall}
  % \vskip -2ex
\end{figure*} 

To bridge the gap between conventional MOT methods and the unique demands of panoramic vision, we introduce \textbf{OmniTrack} (see Fig.~\ref{fig:paradigm}(c))—the first unified MOT framework specifically tailored for 360{\textdegree} panoramic imagery. Unlike standard approaches, OmniTrack is designed to accommodate the challenges inherent to panoramic inputs, \eg,  difficulties in target localization, identity association, and the complex motion dynamics and occlusions unique to wide-angle fields of view.
While End-To-End (E2E, see Fig.~\ref{fig:paradigm}(a)) tracking offers tight integration between detection and association ~\cite{zeng2022motr,zhang2023motrv2,ding2025adatrack++}, it often struggles with the extreme motion dynamics and frequent emergence of new targets in panoramic sequences. 
Conversely, Tracking-By-Detection (TBD, see Fig.~\ref{fig:paradigm}(b)) is more robust to such target variations~\cite{zhang2022bytetrack,yang2024hybrid}, yet it struggles to maintain stable associations for small or distant objects under wide-FoV panoramic views, where targets occupy limited pixels and often suffer from inaccurate distance estimation due to severe geometric distortions.
Motivated by the complementary strengths of both paradigms, OmniTrack unifies E2E and TBD tracking within a single adaptive framework. In panoramic scenarios, rapid motion—particularly of small targets—often yields low IoU across frames, leading to association failure. The E2E paradigm mitigates this by exploiting instance-level features for motion-robust associations, whereas the TBD paradigm, relying on bounding-box matching, remains more resilient to occlusion and overlap. By dynamically switching between the two, OmniTrack effectively handles both high-speed motion and frequent occlusions—key challenges in panoramic multi-object tracking.

Building upon this unified architecture, we further develop \textbf{OmniTrack++}, which addresses the limitations of short-term association in panoramic MOT. While the original OmniTrack demonstrated strong performance, it remained vulnerable to identity drift under prolonged occlusions or significant appearance changes. To overcome these challenges, OmniTrack++ introduces an \emph{ExpertTrack Memory} module that augments long-range trajectory representations, enabling more reliable recovery of fragmented tracks and preserving identity consistency over extended temporal spans. In addition, we refine the \emph{Tracklet Management} mechanism to support adaptive paradigm selection between E2E and TBD modes, allowing the framework to dynamically balance efficiency and robustness according to scene conditions. Together, these enhancements substantially improve the stability and accuracy of panoramic multi-object tracking.

Similar to its predecessor, \textbf{OmniTrack++} continues to center around a unified feedback mechanism that reinjects intermediate tracking cues from preceding frames to guide subsequent predictions. This mechanism serves as the backbone of temporal reasoning, where historical spatial and appearance information continuously informs the decoding process. By integrating these trajectory-aware signals into the attention pathway, the model effectively maintains long-term identity consistency and suppresses temporal drift, even under severe occlusions or rapid viewpoint changes. The feedback loop thus functions not only as a source of robust identity anchors but also as a temporal regularizer that progressively refines the feature space over time. Through this iterative refinement, OmniTrack++ establishes a closed-loop perception framework in which spatial alignment, temporal coherence, and memory-based context jointly enhance the stability and continuity of panoramic multi-object tracking.

Within this framework, four interdependent components collaboratively address the challenges of panoramic multi-object tracking: the \emph{DynamicSSM Block}, the \emph{FlexiTrack Instance}, the \emph{ExpertTrack Memory}, and the \emph{Tracklet Management} module.
The DynamicSSM Block first corrects FoV-induced geometric distortions and establishes a consistent appearance representation across wide-angle views.
Built upon these stabilized features, the FlexiTrack Instance leverages temporal coherence to refine spatial attention and ensure reliable short-term associations.
The ExpertTrack Memory then consolidates these instance representations over extended time spans, providing long-range identity cues that reinforce both localization accuracy and appearance consistency.
Finally, the Tracklet Management module integrates these cues within an adaptive association strategy, dynamically selecting between the E2E and TBD paradigms to maintain trajectory continuity under diverse motion and visibility conditions.
By toggling the data association process within this module, OmniTrack++ can seamlessly transition between the two modes—disabling association yields an E2E variant (OmniTrack++$_{E2E}$), whereas enabling it activates a traditional TBD variant (OmniTrack++$_{DA}$)—as illustrated in Fig.~\ref{fig:paradigm}(c).
Together, these components form a coherent feedback-driven architecture in which geometric correction, temporal modeling, long-term memory, and adaptive paradigm switching jointly enhance the robustness of panoramic MOT.

To facilitate research in panoramic multi-object tracking, we establish the EmboTrack benchmark, a large-scale real-world dataset designed to capture the challenges of dynamic mobile perception under $360^{\circ}$ fields of view (see Fig.~\ref{fig:overall}(a)–(b)).
EmboTrack spans five campuses across two metropolitan regions and comprises $44$ panoramic sequences—totaling $26,400$ annotated frames—representing a $37.5\%$ increase over our previous QuadTrack dataset~\cite{luo2025omnidirectional}.
It consists of two complementary subsets: QuadTrack, collected with a quadrupedal robot equipped with a $360^{\circ}\times70^{\circ}$ FoV panoramic camera, and BipTrack, newly introduced using a bipedal wheel–legged robot fitted with an Insta360 X5 panoramic camera (supporting a ${\sim}170^{\circ}$ single-lens FoV).
The quadruped’s biomimetic gait and the wheel–legged robot’s pitch-induced oscillations jointly produce diverse and realistic motion perturbations, yielding complex non-linear trajectories and frequent viewpoint shifts.
Unlike conventional MOT datasets~\cite{milan2016mot16, caesar2020nuscenes, dendorfer2020mot20, semantickitti, bdd100k, cui2023sportsmot, kondo2025mva}, which typically rely on static or uniformly moving sensors, EmboTrack captures full-surround panoramic imagery under dynamic real-world conditions, providing a unique and challenging benchmark for 360{\textdegree} MOT research.

Extensive experiments on the proposed EmboTrack benchmark validate the effectiveness of \textbf{OmniTrack++}, particularly on the challenging \emph{QuadTrack} subset (see Fig.~\ref{fig:overall}(c)).
In the E2E paradigm, OmniTrack++ achieves an HOTA of $34.90$ and an IDF1 of $41.21$, corresponding to relative improvements of $43\%$ and $52\%$, respectively, over the original OmniTrack.
Under the TBD paradigm, the model attains an HOTA of $36.08$ and an IDF1 of $42.76$, representing $35\%$ and $38\%$ gains.
These results highlight the remarkable effectiveness of the \emph{ExpertTrack Memory} module on datasets with long, continuous trajectories such as QuadTrack, demonstrating its ability to integrate long-term trajectory information into the feedback loop and maintain temporal coherence across extended sequences.
Moreover, the refined \emph{Tracklet Management} module, which adaptively fuses the advantages of both E2E and TBD paradigms, provides more reliable association and improved robustness under complex motion dynamics and frequent vertical oscillations.
Overall, these findings confirm that incorporating long-term trajectory-informed feedback and adaptive paradigm management enables OmniTrack++ to deliver state-of-the-art performance in 360{\textdegree} panoramic MOT.

This journal submission significantly extends upon our preliminary work presented at CVPR 2025~\cite{luo2025omnidirectional}, with key advancements in methodology, benchmark dataset, and experimental verification:
\begin{compactitem}
    \item[(1)] \textbf{ExpertTrack Memory:} We introduce a trajectory-shared Mixture-of-Experts (MoE) memory module that enhances the discriminative power of instance-level features. This design enables more reliable identity preservation and supports robust and consistent trajectory recovery under occlusion, re-entry, or brief target loss.

    \item[(2)] \textbf{Refined Tracklets Management:} We propose a more flexible tracklet management mechanism that enables fine-grained and adaptive paradigm switching between End-To-End (E2E) and Tracking-By-Detection (TBD) modes. This dynamic integration enhances adaptability across tracking conditions and scene complexities.
    
    \item[(3)] \textbf{BipTrack Dataset:} We enrich the established dataset with newly collected sequences captured by a bipedal wheel-legged robot equipped with an Insta360 panoramic camera. Unlike the normally smooth wheeled motion, the robot's locomotion introduces pitch oscillations and gait-like fluctuations, resulting in irregular trajectories that substantially enhance motion diversity. This expansion not only increases the difficulty of motion modeling but also enables research into pre-adaptation, domain generalization, and transfer learning for robotic tracking.
    
    \item[(4)] \textbf{Enriched Experiments and Analyses:} We conduct additional comparative experiments and comprehensive ablation studies to rigorously verify the effectiveness of each proposed component, thereby enhancing both the technical depth, empirical validation, and the overall completeness and reproducibility of the paper.

    \item[(5)] A more detailed description of the proposed methods and other extensions, such as related work discussions and additional qualitative panoramic tracking result analyses.

\end{compactitem}

%% file: Sec_Related_Work.tex
\subsection{Panoramic Scene Understanding}
Panoramic perception enables a holistic understanding of a 360{\textdegree} scene by capturing the entire surrounding environment in a single observation. 
Modern systems achieve this using specialized imaging setups, such as spherical cameras, multi-lens rigs, or ultra-wide field-of-view lenses, which allow seamless acquisition of the full scene~\cite{chen2024360+,dong2024panocontext,ehsanpour2022jrdb_act,jiang2024minimalist}. 
These panoramic inputs preserve spatial continuity across the visual field and provide dense contextual information, supporting downstream perception tasks such as object recognition~\cite{xu2022pandora,dong2024panocontext}, action understanding~\cite{lee2024spatio}, and multimodal scene analysis~\cite{Chen_2024_CVPR,zhu2023unified}.
Moreover, recent works leverage advanced neural architectures, including transformer-based models and Point Spread Function (PSF) aware imaging techniques, to mitigate optical distortions and enhance feature representation across the entire panoramic view~\cite{dong2024panocontext,jiang2024minimalist}.
Current mainstream areas of panoramic perception include panoramic scene segmentation~\cite{zheng2024360sfuda++,yan2023panovos,zhang2024goodsam,zhong2025omnisam,jiang2025multi}, 
panoramic depth estimation~\cite{bai2024glpanodepth,ai2024elite360d,wang2022bifuse++,shen2022panoformer,chang2023depth_neural,zhuang2023spdet}, 
panoramic layout estimation~\cite{yu2023panelnet,shen2023disentangling,ling2023panoswin,shen2024_360_layout}, 
panoramic scene generation~\cite{zhou2025dreamscene360,wang2024360dvd,li2023panogen,ye2024diffpano}, 
panoramic reconstruction~\cite{wang2024perf,fu2023panopticnerf,ren2025panosplatt3r},
and panoramic flow estimation~\cite{li2022deep,liu2025prior_flow}, 
as well as other dense and spatiotemporal understanding~\cite{park2024fully,kim2024fully,fan2024learned,hong2023_par2net,chen2024saliency,cokelek2025spherical}.

Researchers typically unfold panoramas into equirectangular or polyhedral projections to adapt algorithms originally designed for narrow-FoV imagery~\cite{jiang2021unifuse,wang2022bifuse++,li2022deep}. 
However, such projections introduce latitude-dependent distortions and seam artifacts, motivating projection-aware modeling. 
To mitigate these issues, topology- and geometry-aware strategies enforce cyclic continuity at the $0/2\pi$ boundary via wrap-around padding or periodic/spherical attention~\cite{benny2025sphereuformer,wang2024360dvd}, design distortion-aware patch embeddings to improve equirectangular projection (ERP) tokenization~\cite{zhang2024behind,li2023sgat4pass,shen2022panoformer}, inject explicit geometric priors through ERP-aware attention~\cite{yun2023egformer,cao2024geometric} or spherical harmonics~\cite{lee2025hush}, and re-discretize the sphere with icosahedral or Hierarchical Equal Area isoLatitude Pixelization (HEALPix) sampling to obtain near-uniform, rotation-friendly representations~\cite{carlsson2024heal,ai2024elite360d}, as well as polyhedral and spherical formulations such as SpherePHD~\cite{lee2020spherephd} and spherical convolutions~\cite{su2021learning_spherical}, which directly model non-Euclidean geometry on the sphere. 
Deformable operators further alleviate localized projection distortion in high-latitude regions~\cite{shi2023panoflow,zhang2024behind}. 
Complementing these structural remedies, State Space Models (SSMs) provide lightweight global modeling that improves long-range spatial continuity, as shown for panoramic semantic segmentation with vision Mamba~\cite{xu2025mamba4pass} and single-image bird’s-eye-view mapping~\cite{Wei_2024_ACCV}. 
Collectively, projection-aware design, spherical structural priors, and efficient sequence modeling form a coherent toolkit for panoramic scene understanding.

Recent advances in embodied intelligence have increasingly leveraged panoramic visual perception to enable robots to perceive and interact with their surroundings more effectively. 
For example, HumanoidPano~\cite{zhang2025humanoidpano}, Humanoid Occupancy~\cite{cui2025humanoid}, and RobotPan~\cite{ma2026robotpan} integrate spherical panoramic imagery with LiDAR for multimodal perception on humanoid robots, whereas Avatar360~\cite{chalmers2024avatar360}, EmbodiedPlace~\cite{liu2025embodiedplace}, and enhanced language-guided navigation~\cite{wang2024enhanced_language_navigation} exploit panoramic inputs for 6-DoF perception, place recognition, and cross-modal navigation. 
Despite these efforts, the study of MOT within panoramic embodied contexts remains limited, even though MOT is crucial for mobile agents to maintain consistent awareness of multiple dynamic objects, support long-term interaction, and enable informed decision-making in complex environments.

Omnidirectional images offer clear advantages for tracking, maintaining continuous observation without the out-of-view issues inherent to limited field-of-view cameras~\cite{neto2025person_tracking,yang2026robust_oanoramic,chen2026ormot}. 
Early works such as Jiang~\etal~\cite{jiang2021500} propose a $500$FPS omnidirectional tracking system using a three-axis active vision mechanism for fast-moving objects. 
Benchmarks like 360VOT~\cite{huang2023360vot} and 360Loc~\cite{huang2024360loc} provide datasets and evaluation protocols addressing spherical distortions, object localization, and cross-device challenges. 
Xu~\etal~\cite{xu2024360vots} further introduce an extended bounding FoV (eBFoV) representation to mitigate panoramic distortions in videos. 
More recent panoramic tracking datasets and methods, including Leader360V~\cite{zhang2025leader360v}, POT~\cite{pei2021pot}, MMPAT~\cite{he2021know_your_surroundings}, and TPT-Bench~\cite{ye2025tpt_bench}, continue to advance the field, whereas works like Cao~\etal~\cite{cao2025siamese} explore robust tracking for quadrupedal robots, albeit limited to single-object scenarios.

In contrast to earlier studies, we explicitly address the instability of panoramic imagery under real-world dynamic perturbations, including vertical vibrations and pitch fluctuations, and demonstrate that our method robustly supports accurate object localization and temporal association in full-surround environments

\subsection{Multi-Object Tracking}
Tracking-By-Detection (TBD)~\cite{Chen_2024_CVPR,Du_2024_CVPR,qin2024towards,nettrack2024cvpr,huang2024deconfusetrack,lv2024diffmot,li2023ovtrack,qin2023motiontrack} is currently one of the most dominant paradigms in image-based MOT. 
In this framework, object locations are first predicted by a detector and then associated across frames through data association. The paradigm was initially introduced by SORT~\cite{bewley2016simple}, and later extended by DeepSORT~\cite{wojke2017simple}, which incorporated deep appearance features to improve association accuracy and established the foundation of this classic approach. Building upon it, ByteTrack~\cite{zhang2022bytetrack} proposed a confidence-based, stage-wise association strategy to handle low-confidence detections. 
StrongSORT~\cite{du2023strongsort} introduced a keypoint-guided matching method to reduce trajectory fragmentation, while OC-SORT leveraged reliable detectors to predict motion centered on detections, aiding track recovery after target loss. 
UCMCTrack~\cite{yi2024ucmctrack} further addressed the inconsistency between the image plane and the actual motion plane by modeling object movement in the physical space, thereby improving motion prediction. Around the same time, Hybrid-
SORT~\cite{yang2024hybrid} incorporated auxiliary cues such as detection confidence and velocity direction to enhance track management under ambiguous conditions. In addition, TrackTrack~\cite{tracktrack_2025_CVPR} approached the association problem from a tracking-driven perspective and achieved promising results through a global optimization-based refinement strategy, which effectively improved identity consistency and reduced fragmented trajectories in complex scenarios.

\noindent \textbf{End-To-End (E2E)} tracking has recently gained increasing attention in the research community as a promising paradigm that tightly couples detection and tracking within a unified model, eliminating the need for complex data association and post-processing. TransTrack~\cite{transtrack} was among the first to introduce attention mechanisms into the MOT task, using interactions between target and global features to achieve joint detection and tracking. TrackFormer~\cite{meinhardt2021trackformer} further advanced this idea by proposing query-based object localization and association, resulting in a more concise and intuitive design. 
MOTR~\cite{zeng2022motr} extended this approach by incorporating temporal feature aggregation and tailored training strategies to enhance tracking performance. 
MOTRv2~\cite{zhang2023motrv2} focused on improving the detection quality to further boost overall tracking accuracy. 
More recently, MeMOTR~\cite{MeMOTR} introduced a memory bank and query interaction module to enhance feature discriminability across trajectories, reducing identity switches and improving association robustness. 

Unlike existing methods~\cite{gao2025multiple,yan2025comot} that focus on narrow-FoV pinhole camera data with linear sensor motion, we address the challenges of MOT in panoramic-FoV scenarios. 
Specifically, our framework mitigates geometric distortions through a DynamicSSM Block and exploits the full 360{\textdegree} context via trajectory-informed feedback with ExpertTrack Memory, thereby enabling accurate and consistent identity association under panoramic-FoV complex real-world environments.

%% file: Sec_Method.tex
In this section, we introduce OmniTrack++, a panoramic multi-object tracking framework designed to tackle the unique challenges of panoramic FoV images. These include enlarged search spaces, geometric distortions, resolution degradation, and lighting inconsistencies, all of which hinder reliable localization and identity association in MOT.

OmniTrack++ adopts a feedback-driven architecture (Sec.~\ref{subsec:Framework}) that iteratively refines object detection by reintegrating trajectory information into the perception pipeline.
This design enhances tracking accuracy and consistency under panoramic FoV conditions.
Within this framework, four interdependent modules collaboratively address the major challenges of panoramic multi-object tracking (Fig.~\ref{fig:paradigm}(c)):

\begin{compactitem}
    \item \textbf{DynamicSSM Block} (Sec.~\ref{subsec:CSEM}): Serves as the geometric foundation of the framework by mitigating panoramic distortions and photometric inconsistencies. It stabilizes spatial feature representations across the $360^{\circ}$ view, providing reliable inputs for subsequent temporal modeling.
    
    \item \textbf{FlexiTrack Instance} (Sec.~\ref{subsec:Temporal_Query}): builds upon stabilized features from the DynamicSSM Block to establish short-term temporal coherence. Leveraging trajectory-informed feedback, it refines attention over the panoramic scene and supports precise localization and identity association.
    
    \item \textbf{ExpertTrack Memory} (Sec.~\ref{subsec:ExpertTrack}): Extends FlexiTrack’s short-term reasoning to long-term identity preservation. It consolidates appearance features via a combination of stable and dynamic memories, aided by a shared MoE that captures diverse distortions and illumination changes typically found in panoramic imagery.
    
    \item \textbf{Tracklet Management} (Sec.~\ref{subsec:Tracklets_Management}): Integrates information from preceding modules to maintain trajectory continuity. It adaptively toggles between E2E and TBD paradigms—disabling association yields an E2E variant (OmniTrack++${_{E2E}}$), while enabling it activates a TBD variant (OmniTrack++${_{DA}}$)—thus balancing efficiency and robustness under varying scene dynamics.

\end{compactitem}

Together, these four modules form a coherent feedback-driven system in which geometric correction, temporal modeling, long-term memory, and adaptive association operate in concert to achieve consistent, distortion-resilient, and identity-stable tracking performance, particularly when navigating through complex panoramic environments.

\subsection{Feedback Mechanism}
\label{subsec:Framework}

\input{algorithm}

\begin{figure}[!t]
  \centering
  \includegraphics[width=0.49\textwidth]{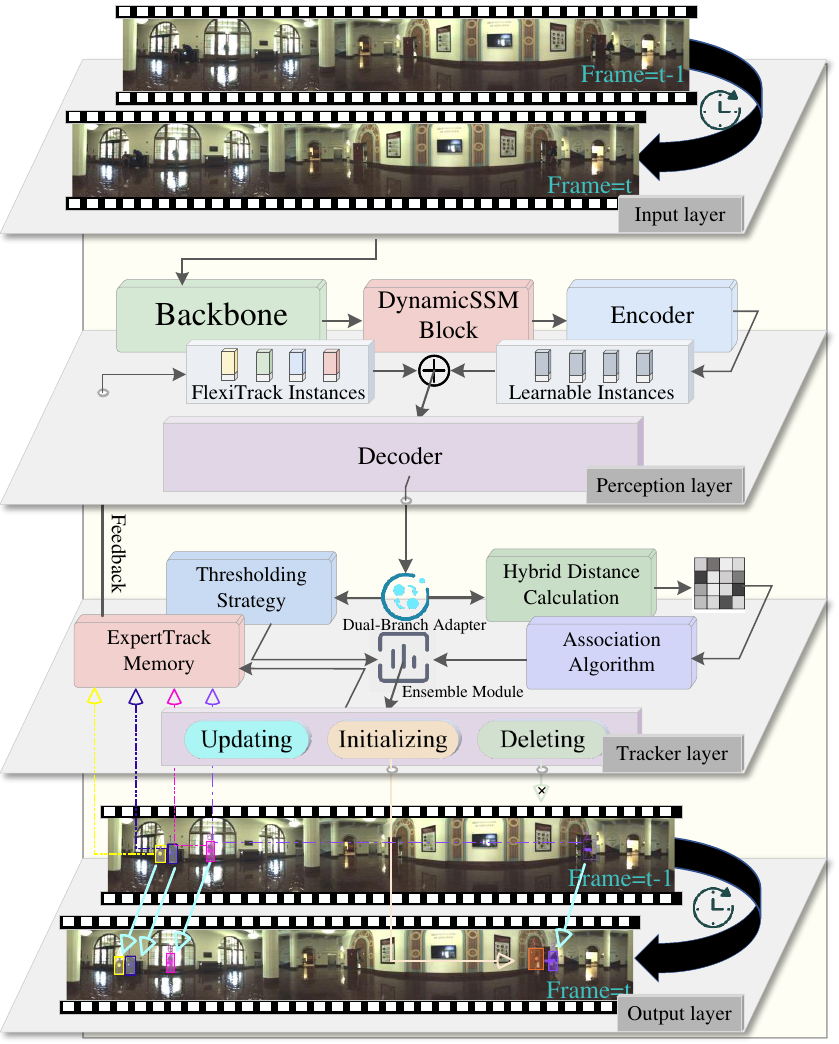}
      \caption{
    Pipeline overview of OmniTrack++. 
    At frame $t$, the panoramic input is processed by a shared backbone, a DynamicSSM block, and an encoder to produce \emph{learnable instances} for the current frame. 
    In parallel, \emph{FlexiTrack Instances} from frame $t-1$ are retrieved from the \emph{ExpertTrack Memory}. These two sets of tokens are concatenated and fed into the decoder to generate object proposals. 
    A Dual-Branch Adapter then routes them to either (i) a TBD branch, using hybrid distance calculation and an association algorithm for trajectory updates, or (ii) an E2E branch, using a thresholding strategy for direct updates. An Ensemble Module fuses both outputs to yield the final track set, which is written back to the ExpertTrack Memory to instantiate the FlexiTrack Instances for frame $t+1$, closing the feedback loop.}
    
      \label{fig:pipeline}
  % \vskip -2ex
\end{figure}

The OmniTrack++ framework (Fig.~\ref{fig:pipeline}) incorporates a dedicated feedback mechanism that refines detections by propagating trajectory information from previous frames back into the detector. This design is motivated by the high uncertainty inherent in panoramic imagery, where severe distortions, wide-FoV dynamics, and frequent target re-appearance often destabilize detection and association. By leveraging trajectory-informed feedback, OmniTrack++ progressively constrains uncertainty and enhances the stability of multi-object tracking in panoramic fields of view.

In conventional MOT pipelines~\cite{zhang2022bytetrack,cao2023observation}, detection and association are decoupled, and each frame is processed independently. The entropy of detections at frame $t$ can be expressed as
\begin{align}
H(x_t) = -\sum_{i=1}^{n} P(x_t^i) \log P(x_t^i),
\end{align}
where $x_t^i$ denotes the position of the $i$-th target with probability distribution $P(x_t^i)$. When extended across $T$ frames, the cumulative entropy of independent matching is formulated as
\begin{align}
H_{\text{ind}} = \sum_{t=1}^{T} H(x_t) + H(\{y_t\}),
\end{align}
where $H(\{y_t\})$ represents the entropy of global association over trajectories $\{y_t\}$.

In contrast, OmniTrack++ introduces feedback such that detections at frame $t-1$ guide those at frame $t$. The conditional entropy of detections at frame $t$, given prior feedback $y_{t-1}$, is expressed as
\begin{align}
H(x_t | y_{t-1}) = -\sum_{i=1}^{n} P(x_t^i | y_{t-1}^i) \log P(x_t^i | y_{t-1}^i).
\end{align}
The resulting total entropy is
\begin{align}
H_{\text{fb}} = \sum_{t=1}^{T} H(x_t | y_{t-1}),
\end{align}
which satisfies $H_{\text{fb}} < H_{\text{ind}}$, indicating a consistent reduction in uncertainty over time. This feedback-driven conditioning further constrains uncertainty, supporting precise localization and accurate association in panoramic-FoV scenarios.

\subsection{DynamicSSM Block} 
\label{subsec:CSEM}

\begin{figure}[!t]
    \centering
      \includegraphics[width=0.48\textwidth]{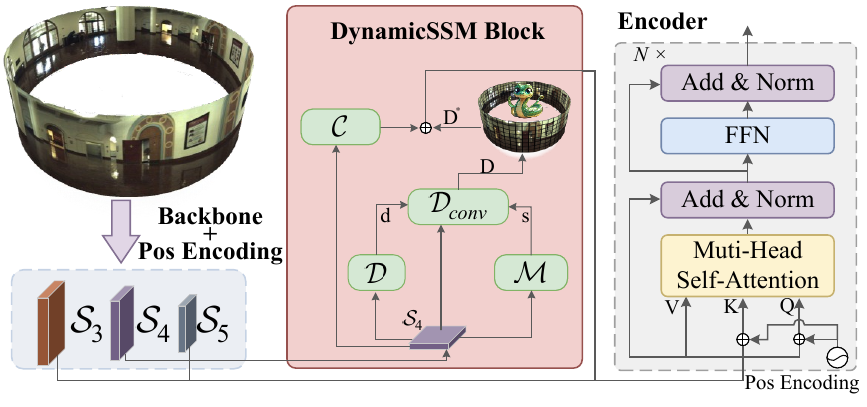}
      \caption{The proposed DynamicSSM Block is integrated into a standard DAB encoder as a plug-in enhancement. Rather than explicitly modeling panoramic geometry, it implicitly calibrates spatial and photometric feature distributions to mitigate geometric distortions and illumination variation. This adaptation yields more robust and stable representations, enabling more reliable decoding and multi-object tracking in panoramic scenes.}
      \label{fig: CircularStatE Module}
\end{figure}

While FlexiTrack Instances effectively encode trajectory-informed feedback, panoramic imagery remains prone to geometric distortion and photometric inconsistency, especially under wide-FoV and high-dynamic-range conditions. These artifacts degrade instance stability and hinder reliable association. To address this, we introduce a \textbf{DynamicSSM Block} (see Fig.~\ref{fig: CircularStatE Module}) to refine features by mitigating distortion and enhancing photometric consistency.

In our earlier OmniTrack framework, the DynamicSSM Block was integrated with the RT-DETR~\cite{Zhao_2024_CVPR} encoder to form the CircularStatE Module. In OmniTrack++, we further adapt this block into the DAB Transformer encoder~\cite{zhu2020deformable}, where it operates as a plug-in enhancement. This adaptation enables the block to implicitly calibrate spatial and photometric distributions without explicitly modeling panoramic geometry, thereby producing more stable and robust features for downstream decoding and tracking.

Formally, the DynamicSSM Block operates in four stages:

\noindent (1) Distortion and Scale Estimation.  
Given the input feature map $\mathbf{S}_4$, a distortion field $\mathbf{D}$ and a scale prior $\mathbf{S}$ are predicted via two lightweight modules:
\begin{align}
\mathbf{D},\, \mathbf{S} = \mathcal{D}_{est}(\mathbf{S}_4),\, \sigma(\mathcal{S}_{est}(\mathbf{S}_4)),
\end{align}
where $\mathbf{D}, \mathbf{S} \in \mathbb{R}^{B \times C \times H \times W}$ encode geometric deformation and spatial scale, and $\sigma(\cdot)$ denotes the activation function.

\noindent (2) Distortion-Aware Refinement.  
A dynamic convolution module adaptively modulates features based on the predicted cues:
\begin{align}
\mathbf{Z} = \mathcal{D}_{conv}(\mathbf{D} \odot \mathbf{S},\, \mathbf{S}_4),
\label{eq:dynamic_conv}
\end{align}
where $\odot$ denotes element-wise modulation.

\noindent (3) Long-Range Consistency via State Space Model.  
To enhance photometric stability, we apply a multi-directional SSM~\cite{mamba2}:
\begin{align}
\mathbf{Z}^{\star} = \frac{1}{L}\sum_{d=1}^{L} F_{S6}(\mathcal{S}_{d}(\mathbf{Z})),
\label{eq:ssm}
\end{align}
where $\mathcal{S}_{d}$ is the directional scanning operator, $L$ the number of directions, and $F_{S6}$ denotes the S6 transformation. 

\noindent (4) Feature Fusion.  
The refined representation is fused with a residual CNN branch:
\begin{align}
\mathbf{F} = \mathcal{F}\!\left( \mathcal{C}(\mathbf{S}_4) \oplus \mathbf{Z}^{\star} \right),
\label{eq:fusion}
\end{align}
where $\oplus$ denotes feature fusion. The final output $\mathbf{F}$ provides distortion-compensated and photometrically stabilized features.

By embedding this plug-in into the DAB Transformer encoder, OmniTrack++ benefits from more reliable feature encoding tailored to panoramic imagery. In synergy with FlexiTrack Instances, the DynamicSSM Block ensures that both geometric and temporal uncertainties are effectively reduced, yielding stable representations that support precise localization and accurate association across challenging wide-FoV tracking scenarios.

\subsection{FlexiTrack Instance}
\label{subsec:Temporal_Query}

While the Tracklets Management module governs paradigm switching and trajectory lifecycle, the realization of trajectory-informed feedback is achieved through the proposed FlexiTrack Instance. In OmniTrack++, feedback information is encoded into FlexiTrack Instances, which are injected with standard Learnable Instances into the shared decoder (see the \emph{Perception Layer} in Fig.~\ref{fig:pipeline}). This allows the decoder to jointly process detection-driven and trajectory-informed representations, enabling precise localization and temporal association without exhaustive search across the panoramic FoV.

Each FlexiTrack Instance inherits its structure from the Learnable Instance, consisting of a feature vector $\mathcal{X} \in \mathbb{R}^{c_s}$ and an anchor $\mathcal{Y} \in \mathbb{R}^{c_s}$, both residing in a compact embedding space. During training, stochastic perturbations are applied to both components,
\begin{align}
\mathcal{X}' = \mathcal{X} + \mathcal{N}_X, \quad
\mathcal{Y}' = \mathcal{Y} + \mathcal{N}_Y,
\end{align}
where $\mathcal{N}_X$ and $\mathcal{N}_Y$ denote additive noise terms.
This regularization mitigates over-reliance on historical cues and improves generalization to unseen trajectories.
For initialization, let $\mathcal{I_F}$ denote the set of FlexiTrack Instances corresponding to $N$ trajectories:
\begin{align}
\mathcal{I_F} = { \mathcal{I_F}^i \mid
\mathcal{I_F}^i = (\mathcal{X}'_i, \mathcal{Y}'_i), ,
i \in {1,2,\dots,N} },
\end{align}
where $\mathcal{X}'_i, \mathcal{Y}'_i \in \mathbb{R}^{c_s}$ represent the trajectory-informed feature vector and anchor of the $i$-th tracklet, $c_s$ is the feature dimension.

By embedding trajectory knowledge into a decoder-compatible instance form, FlexiTrack Instances serve as the operational vehicle of the feedback mechanism. They guide the decoder’s attention toward relevant spatial regions, reduce ambiguity in temporal association, and enable seamless integration with both E2E and TBD paradigms. This design provides the flexibility to unify diverse tracking strategies while directly exploiting trajectory-informed cues, thereby improving localization accuracy and association stability in challenging panoramic wide-FoV scenarios.

\subsection{ExpertTrack Memory}
\label{subsec:ExpertTrack}
\begin{figure*}[htb]
  \centering
  \includegraphics[width=0.98\textwidth]{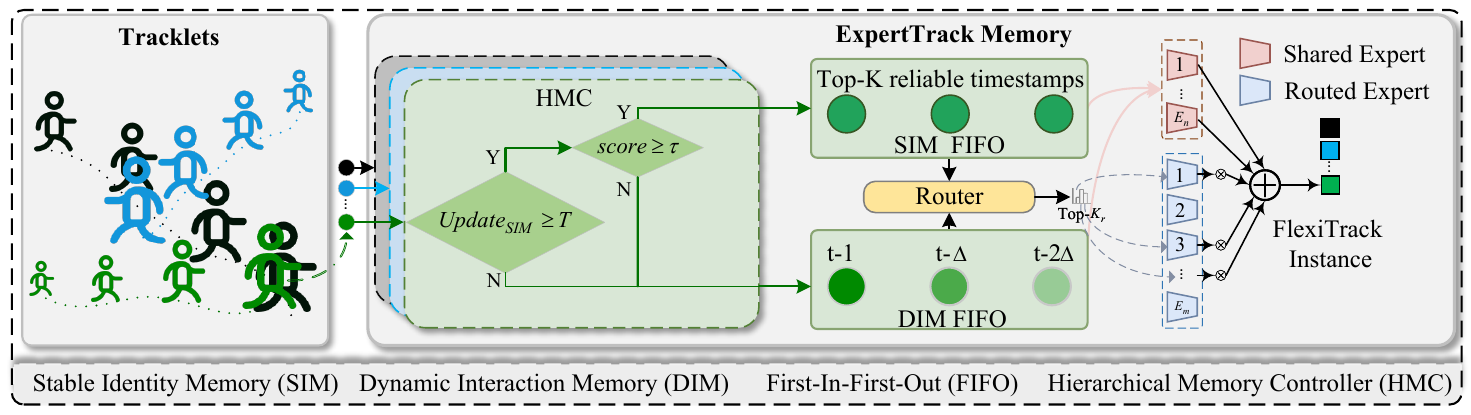}
  \caption{ExpertTrack Memory framework. 
The module integrates long-term Stable Identity Memory (SIM) and short-term Dynamic Interaction Memory (DIM) to jointly maintain identity consistency and adapt to rapid appearance changes under panoramic distortions. 
A Hierarchical Memory Controller (HMC) assigns high-confidence features to SIM and recent-frame updates to DIM. 
A Router then selects the top-$K_r$ features across both memories and forwards them to a Shared Mixture-of-Experts (MoE) module, where specialized experts handle diverse appearance variations—such as illumination inconsistency and geometric deformation. 
The aggregated expert outputs are fused into the FlexiTrack Instance, enabling robust and adaptive identity association across panoramic views.}
  \label{fig:moe}
\end{figure*}

Although the DynamicSSM Block alleviates geometric and photometric inconsistencies, reliable long-term identity preservation in panoramic multi-object tracking remains highly challenging.
Wide-FoV distortions, frequent occlusions, and severe appearance shifts often cause embedding drift, particularly when targets reappear after long absences or under drastic viewpoint changes.
To overcome these limitations, we propose the ExpertTrack Memory~(Fig.~\ref{fig:moe}), a hierarchical memory-driven module that unifies long-term identity stability with adaptive appearance modeling under panoramic distortions.

Hierarchical Memory Organization.
Each instance $i$ is equipped with a compact memory bank $\mathcal{M}_i$ storing representative embeddings:
\begin{equation}
\mathcal{M}_i = \{ \mathbf{f}_i^{(t_1)}, \mathbf{f}_i^{(t_2)}, \dots, \mathbf{f}_i^{(t_{n_m})} \}.
\end{equation}
The first half, $\{\mathbf{f}_i^{(t_1)}, \dots, \mathbf{f}_i^{(t_{n_m/2})}\}$, forms the \emph{Stable Identity Memory}, containing confidence-selected keyframes for long-term identity consistency. The second half, $\{\mathbf{f}_i^{(t_{n_m/2+1})}, \dots, \mathbf{f}_i^{(t_{n_m})}\}$, constitutes the \emph{Dynamic Interaction Memory} for capturing short-term appearance and motion variations. This dual structure balances stability and adaptivity, providing a reliable temporal base.

Shared Mixture-of-Experts (MoE).
Panoramic MOT involves highly diverse appearance changes from illumination shifts, geometric distortions, and view-dependent deformations, which challenge a single embedding function.
To address this heterogeneity, we employ a \emph{Shared Mixture-of-Experts (MoE)}~\cite{dai2024deepseekmoe} that dynamically decouples feature adaptation across multiple expert pathways.
Each expert specializes in compensating a particular type of variation—\eg, lighting, orientation, or distortion—while a shared routing mechanism aggregates their responses.
Given a query $q_i$, the attention-based routing yields:
\begin{equation}
\mathbf{f}_i^{\text{sh}} = \sum_{k=1}^{n_e} \alpha_{i,k} \, E_k(\mathbf{q}_i), \quad
\alpha_{i,k} = \frac{\exp(\text{score}(\mathbf{q}_i, E_k))}{\sum_{j=1}^{n_e} \exp(\text{score}(\mathbf{q}_i, E_j))},
\end{equation}
where $E_k(\cdot)$ is the $k$-th expert and $n_e$ the total number of experts.
This design combines specialization and parameter sharing, enhancing adaptability to panoramic distortions.

Expert-Guided Memory Fusion.
To integrate personalized cues with expert-driven adaptation, a gated selector retrieves the  discriminative embedding from $\mathcal{M}_i$ according to query context:
\begin{equation}
\mathbf{f}_i^{\text{pl}} = \text{GatedSelect}(\mathcal{M}_i, \mathbf{q}_i).
\end{equation}
The final embedding merges the personalized retrieval with the shared expert response:
\begin{equation}
\hat{\mathbf{f}}_i = \lambda_i \mathbf{f}_i^{\text{sh}} + (1 - \lambda_i) \mathbf{f}_i^{\text{pl}},
\end{equation}
where $\lambda$ is a learnable balancing parameter.
This joint fusion progressively refines feature embeddings, preserving instance-level distinctiveness while adapting to evolving visual conditions.

By integrating hierarchical memory organization, expert-based adaptation, and gated fusion within a unified feedback framework, the ExpertTrack Memory enables OmniTrack++ to maintain long-term consistency and adaptive resilience under the severe geometric and photometric challenges of panoramic multi-object tracking.

\subsection{Tracklets Management}
\label{subsec:Tracklets_Management}

While the feedback mechanism reduces uncertainty by conditioning detections on historical trajectories, maintaining stable and flexible trajectory evolution requires a dedicated management module. To this end, OmniTrack++ introduces a Tracklets Management component, which serves as the backbone for both paradigm switching and trajectory lifecycle control.

The core of this module is a \emph{Dual-Branch Adapter}, which operates exclusively during inference. Based on the current state of trajectory association, the adapter dynamically selects between the E2E (Alg.~\ref{alg:Omnitrack}, Lines 19-26) and TBD branches (Alg.~\ref{alg:Omnitrack}, Lines 15-17), or ensembles their outputs when complementary advantages can be exploited (Alg.~\ref{alg:Omnitrack}, Line 13). This adaptive switching allows the system to respond to diverse tracking conditions: E2E is favored when motion dynamics are stable and detection quality is high, while TBD provides resilience under target re-entry, partial occlusion, or appearance ambiguity. The ensemble mode further balances these paradigms, mitigating the weaknesses of each when tracking panoramic sequences characterized by large viewpoint shifts and frequent target transitions.

Beyond paradigm selection, Tracklets Management supervises the entire trajectory lifecycle, including initialization, update, and termination. By monitoring instance confidence and temporal consistency, it ensures that tracklets are retained only when supported by sufficient evidence, thereby suppressing spurious associations while preserving long-term identity continuity. During training, this module remains transparent and does not alter gradient flow, but at inference, it functions as a high-level controller that orchestrates the interaction between detection, association, and memory.

Overall, Tracklets Management constitutes a pivotal support module in OmniTrack++, enabling flexible paradigm integration and robust trajectory lifecycle governance. In combination with the feedback mechanism, it provides the structural foundation for precise localization and association across complex panoramic multi-object tracking scenarios.

%% file: Algorithm.tex
\begin{algorithm}
\caption{OmniTrack Inference Process}
\algorithmfootnote{In \textcolor{codegreen}{green} is the key of our method. }

\label{alg:Omnitrack}
\footnotesize
\KwIn{
A Panoramic video/image sequence $\texttt{V}$
}

\KwOut{
Tracks $\mathcal{T}$ of the video/image sequence
}

Initialization: $\mathcal{T} \leftarrow \emptyset$\;

Define the Initialize threshold $\mathcal{\tau}_\mathcal{I}$ \;
Define the Update threshold  $\mathcal{\tau}_\mathcal{U}$ \;

\For{frame $f_k$ in $\texttt{V}$}{
\tcc{As shown in Fig. \ref{fig:pipeline}}

$\{\mathcal{S}_3, \mathcal{S}_4, \mathcal{S}_5\} \leftarrow \texttt{Backbone}(f_k)$ \;

\textcolor{codegreen}{$ \mathcal{I}_L \leftarrow  \texttt{DynamicSSM}(\{\mathcal{S}_3, \mathcal{S}_4, \mathcal{S}_5\})$} \;

\textcolor{codegreen}{$ \mathcal{I}_F \leftarrow \texttt{ExpertTrackMemory}(\mathcal{T}_{f_{k-1}}$}) \;

$ \textcolor{codegreen}{\mathcal{D}_k^F}, \mathcal{D}_k^L \leftarrow  \texttt{Decoder} ( \textcolor{codegreen}{\mathcal{I}_F}, \mathcal{I}_L) $ \;

\BlankLine
\BlankLine
\tcc{Dual-Branch Adapter}

$\mathcal{A} \leftarrow \texttt{BranchController}(\textcolor{codegreen}{{\mathcal{D}_k^F}}+ \mathcal{D}_k^L, \mathcal{T}_{f_{k-1}})$\;

\BlankLine
\BlankLine
\uIf{$\{\text{'TBD'}, \text{'E2E'}\} \subseteq \mathcal{A}$}{
    $\mathcal{T}^{\text{TBD}} \leftarrow \texttt{TBD\_Association}(\textcolor{codegreen}{{\mathcal{D}_k^F}}+ \mathcal{D}_k^L, \mathcal{T}_{f_{k-1}})$\;
    $\mathcal{T}^{\text{E2E}} \leftarrow \texttt{E2E\_Thresholding}(\textcolor{codegreen}{{\mathcal{D}_k^F}}+ \mathcal{D}_k^L, \mathcal{T}_{f_{k-1}})$\;
    \textcolor{codegreen}{$\mathcal{T}_{f_{k}} \leftarrow \texttt{Ensemble}(\mathcal{T}^{\text{TBD}}, \mathcal{T}^{\text{E2E}})$}\;
}
\uElseIf{$\text{'TBD'} \in \mathcal{A}$}{

    \tcc{TBD\_Association}
    $\mathcal{C} \leftarrow  \texttt{DistanceCalculation}(\textcolor{codegreen}{{\mathcal{D}_k^F}}+ \mathcal{D}_k^L, \mathcal{T}_{f_{k-1}})$ \;

    $\{\texttt{Update, Initialize, Delate} \} \leftarrow \texttt{AssociationAlgorithm}(\mathcal{C})$ \;

    $\mathcal{T}_{f_{k}} \leftarrow \{\texttt{Update, Initialize, Delete} \} $
}
\uElseIf{$\text{'E2E'} \in \mathcal{A}$}{

    \tcc{E2E\_Thresholding}
        
            \textcolor{codegreen}{\For{$d$ in $ \{\mathcal{D}_k^F \cup \mathcal{D}_k^L$\} }{
	            \If{$ d \in \mathcal{D}_k^F \And d.score > \mathcal{\tau}_\mathcal{U} $}{
	               $\texttt{Update} \leftarrow  d$ \;
                }
                    \If{$ d \in \mathcal{D}_k^L \And d.score > \mathcal{\tau}_\mathcal{I}  $}{
	               $\texttt{Initialize } \leftarrow  d$ \;
                }
                \Else{
                    $\texttt{Delete} \leftarrow  d$ \;
                }
            }
        }
        $\mathcal{T}_{f_{k}} \leftarrow \{\texttt{Update, Initialize, Delete} \} $
}
\Else{
    \textbf{raise} \texttt{RuntimeError("No valid tracking paradigm activated.")}\;
}
}

\textbf{Return}: $\mathcal{T}$

\end{algorithm}

%% file: Sec_EmboTrack.tex
\label{sec:QuadTrack}
Most existing MOT datasets~\cite{milan2016mot16,dendorfer2020mot20,peize2021dance} are captured with pinhole cameras, which typically feature a narrow FoV and linear sensor motion. However, for panoramic-FoV devices, even slight ego-motion can induce drastic scene changes, introducing substantial challenges for object tracking.

\begin{figure}[t]
    % \vskip -2 ex
    \centering
    \includegraphics[width=\linewidth]{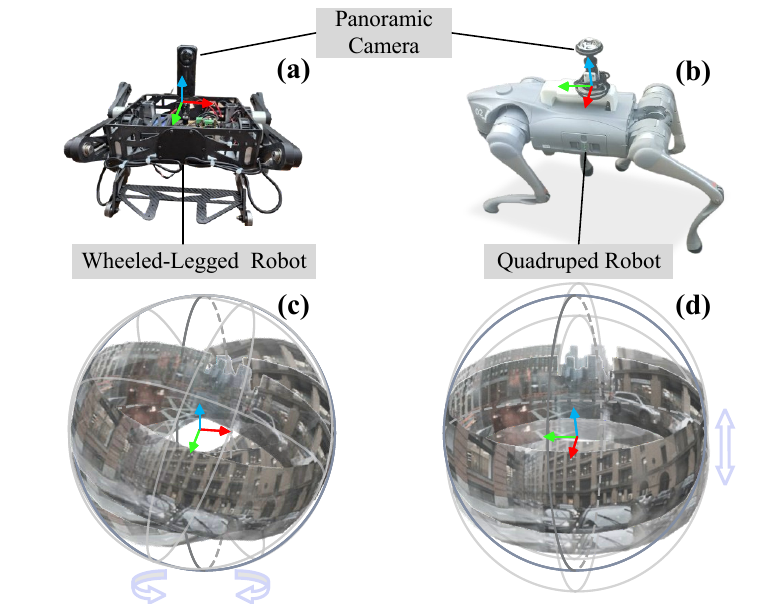}
    \caption{(a) and (b) illustrate the data collection platforms used in our QuadTrack dataset. Specifically, (a) shows a wheeled-legged robot~\wheelleg, whereas (b) depicts a quadrupedal robot~\robotdog. (c) presents the pitch motion noise induced by the movement of the platform in (a), whereas (d) illustrates the vertical (z-axis) oscillation noise generated by the movement of the platform in (b).}
    \label{fig:robot_platform}
    % \vskip -2 ex
\end{figure}

To address this limitation, we introduce EmboTrack, a comprehensive panoramic MOT dataset that significantly extends our previous QuadTrack benchmark. EmboTrack consists of two complementary subsets: (i) BipTrack: Recorded with a wheeled-legged robot equipped with an Insta360 panoramic camera (Fig.~\ref{fig:robot_platform} (a)), it introduces hybrid locomotion dynamics. This platform combines wheel-based mobility with articulated leg joints, producing distinctive motion characteristics such as pitch variations, lateral tilting, and occasional gait-like steps. These motions induce complex scene deformations and non-uniform perspective transitions, complicating object detection and association. (ii) QuadTrack: Collected using a quadruped robotic platform (Fig.~\ref{fig:robot_platform} (b)), it captures panoramic sequences under gait-induced non-linear motion. The quadruped's periodic stepping patterns lead to pronounced pitch and roll fluctuations, vertical body oscillations, and abrupt velocity changes. These dynamics cause frequent viewpoint shifts, unstable camera trajectories, rapid target displacement, partial occlusions, and motion blur, presenting a challenging environment for tracking.
By integrating these two distinct motion paradigms, EmboTrack enhances data diversity, encompassing heterogeneous robotic platforms and complex locomotion patterns. This design enables a more rigorous evaluation of MOT algorithms under panoramic, non-uniform, and dynamically varying conditions, providing a comprehensive benchmark for advancing panoramic multi-object tracking research.

\subsection{QuadTrack: Collection and Characteristics}
To capture panoramic FoV sequences with complex locomotion-induced dynamics, we employ a quadruped robotic platform equipped with a Panoramic Annular Lens (PAL) camera. The quadruped robot was chosen for its biomimetic gait, which closely emulates natural animal locomotion and introduces highly non-linear motion patterns. Such gait-induced oscillations manifest as periodic pitch and roll variations, vertical shaking, and abrupt velocity changes, all of which create challenging conditions for multi-object tracking. The robot’s high maneuverability allows data collection across diverse real-world settings, including sidewalks, open squares, and campus roads, while maintaining stable operation over inclines and obstacles.

The PAL camera provides a $360^\circ{\times}70^\circ$ panoramic FoV at $2048{\times}2048$ resolution and up to $40.5$ FPS, ensuring wide-area scene coverage. Mounted at the top of the quadruped, the camera delivers an unobstructed perspective, enabling panoramic data acquisition in unconstrained outdoor environments. The dataset spans multiple times of day—from morning to evening—across five campuses in the cities of Changsha and Hangzhou, including multiple sites from Hunan University, Central South University, and Hunan Normal University in Changsha, as well as Zhejiang University’s main and Zijingang campuses in Hangzhou, thereby capturing diverse illumination conditions and scene complexities across distinct urban environments.

\begin{figure}[t]
    % \vskip -2 ex
    \centering
    \includegraphics[width=\linewidth]{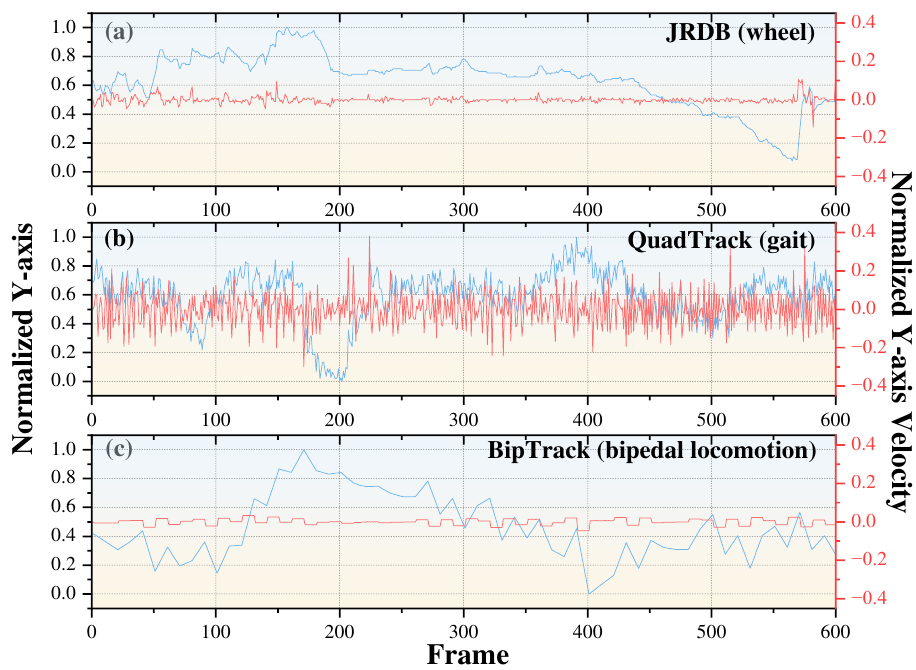}
    \caption{Instance motion trajectories over time. The horizontal axis denotes frames, while the blue curve indicates the normalized $y$-axis position and the red curve represents its temporal variation (\textit{i.e.}, velocity). (a) JRDB dataset captured with a wheeled robotic platform, (b) QuadTrack with a quadrupedal gait platform causing vertical oscillations, and (c) BipTrack with a wheeled–legged platform introducing pitch-induced motion variations.}
    \label{fig:motion_curve}
    % \vskip -2 ex
\end{figure} 

\subsection{BipTrack: Collection and Characteristics}
BipTrack is constructed using a wheeled-legged robotic platform equipped with an \textbf{Insta360 panoramic camera}. Unlike conventional wheeled robots, this hybrid platform integrates wheel-based mobility with articulated leg joints, yielding a unique locomotion style that combines smooth rolling with gait-like perturbations. Such dynamics introduce additional complexity to camera motion, including pitch fluctuations, lateral tilting, and intermittent stepping motions, resulting in panoramic sequences characterized by compound ego-motion and non-uniform viewpoint transitions.

The onboard Insta360 camera provides high-quality panoramic capture with a $360^\circ$ field of view at resolutions up to $3840{\times}1920$ and frame rates of $100$ FPS. With a $1/2$-inch sensor, F1.9 aperture, and $6.7$mm equivalent focal length, the camera is capable of producing detailed wide-angle imagery under varying illumination conditions. Mounted on top of the wheeled-legged robot, the camera maintains a broad perspective while faithfully recording the subtle instabilities induced by hybrid locomotion.

\subsection{EmboTrack: Integrated Benchmark}
EmboTrack integrates both QuadTrack and BipTrack, offering a comprehensive benchmark for multi-object tracking under panoramic video captured from distinct robotic locomotion modalities. By combining the gait-induced oscillations of wheeled-balanced robots with the compound dynamics of wheeled-legged platforms, EmboTrack captures the variability and instability encountered in real-world mobile vision systems, enabling a rigorous evaluation of MOT algorithms.

The dataset spans more than $26,000$ annotated frames of panoramic video, recorded across diverse environmental conditions, object densities, and motion patterns. Each sequence is annotated with consistent identity labels, bounding boxes, and trajectory associations, following a unified annotation protocol. The annotation process employs a hybrid approach that combines automated techniques with manual verification. Initially, bounding boxes and identity labels are generated through an automated linear propagation method based on temporal consistency. These automated annotations are then refined through visual inspection and manual correction to ensure label accuracy. A final quality assurance step involves visualizing the trajectories and making necessary adjustments, ensuring high annotation reliability.

EmboTrack integrates QuadTrack and BipTrack to form a unified panoramic MOT benchmark that captures a broad spectrum of locomotion-induced challenges. It encompasses sequences characterized by frequent pitch oscillations, lateral tilts, and compound motion patterns, which collectively introduce severe viewpoint shifts, oscillatory motion, and distortion effects. These dynamics amplify the difficulties of maintaining consistent object identities and accurate associations. By incorporating such diverse motion characteristics, EmboTrack enables comprehensive evaluation of both TBD and E2E paradigms under realistic robotic locomotion, ensuring balanced assessment across diverse motion conditions.

\input{Table_Dataset_Comparison}

\subsection{Data Distribution and Comparative Analysis}

To contextualize EmboTrack within the broader landscape of  MOT benchmarks, we first compare it with representative datasets (Tab.~\ref{tab:comparison dataset}). 
Traditional MOT datasets such as KITTI~\cite{geiger2013vision}, MOT17~\cite{milan2016mot16}, and MOT20~\cite{dendorfer2020mot20} are primarily captured with pinhole cameras and limited viewpoints, often lacking egocentric motion and thus offering relatively constrained motion diversity. 
Autonomous driving datasets, \eg, Waymo~\cite{waymo}, nuScenes~\cite{caesar2020nuscenes}, and BDD100K~\cite{bdd100k} introduce large-scale driving scenarios, yet they remain dominated by wheeled platforms that yield smooth and predictable trajectories. Internet-sourced datasets such as SportsMOT~\cite{cui2023sportsmot} and DanceTrack~\cite{peize2021dance} shift the focus toward dynamic human activities, but their reliance on curated footage induces biases in scene composition and motion dynamics. JRDB~\cite{martin2021jrdb}, while one of the few benchmarks captured from a mobile robot with panoramic coverage, still reflects the stable motion patterns of wheeled locomotion.

\begin{figure}[!t]
    % \vskip -2 ex
    \centering
      \includegraphics[width=0.48\textwidth]{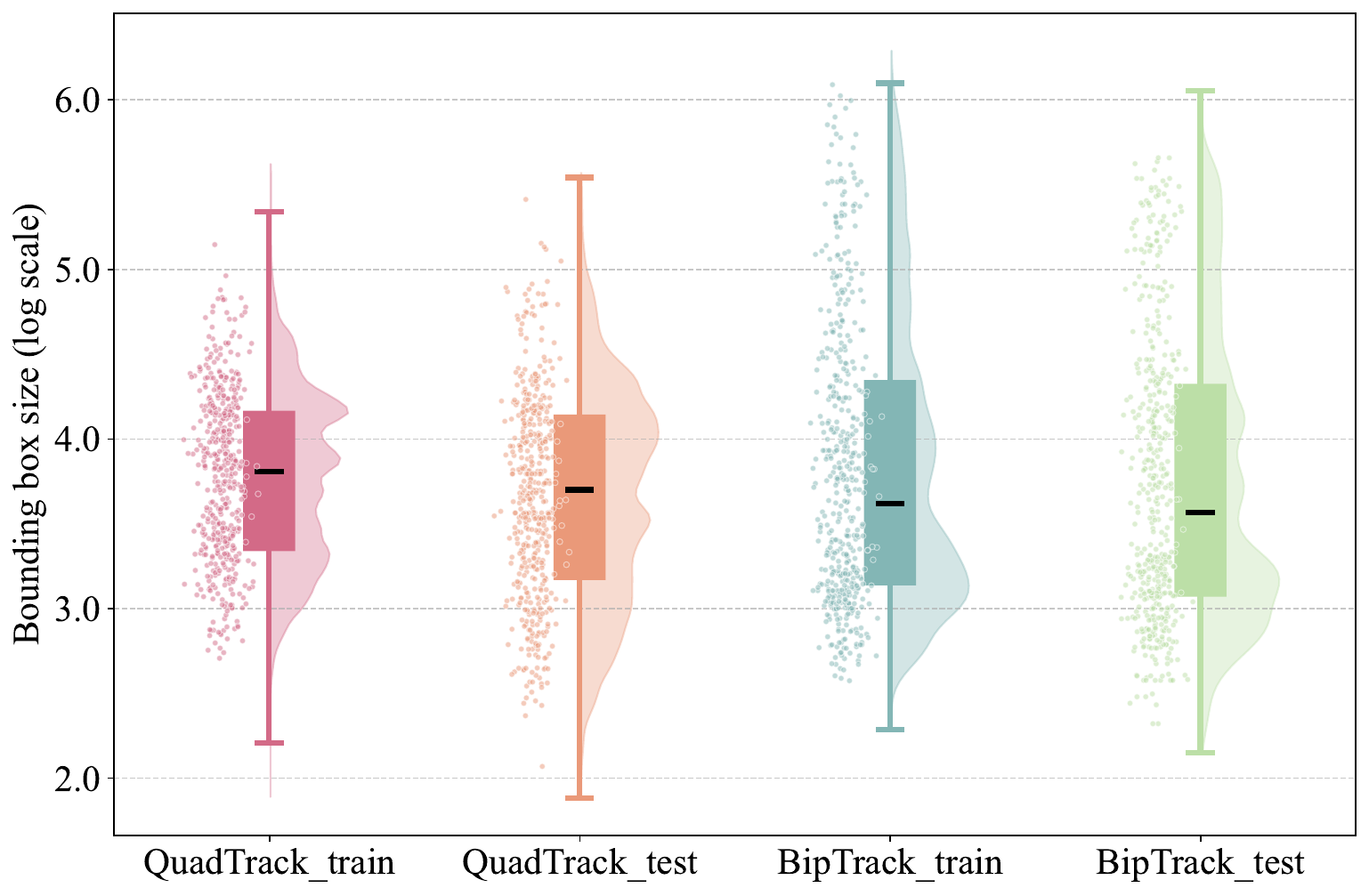}
      % \vskip -2ex
      \caption{Distribution of bounding box sizes in EmboTrack. The plots illustrate the box size distributions of the QuadTrack and BipTrack datasets across both training and test sets.}
      \label{fig:distribution_box}
    % \vskip -2 ex
\end{figure} 

In contrast, EmboTrack introduces a new class of embodied egocentric tracking data characterized by locomotion-induced motion perturbations. Fig.~\ref{fig:motion_curve} shows normalized Y-axis pixel displacements over time in the image plane: JRDB sequences exhibit relatively smooth displacement (Fig.~\ref{fig:motion_curve} (a)), while QuadTrack, recorded on a quadrupedal platform, produces oscillatory shifts due to gait (Fig.~\ref{fig:motion_curve} (b)), with corresponding disturbances shown in Fig.~\ref{fig:robot_platform} (d). BipTrack further extends this paradigm with a hybrid wheeled–legged platform, where otherwise smooth wheeled trajectories are intermittently disturbed by gait-like oscillations and pitch variations, as illustrated in Fig.~\ref{fig:motion_curve} (c) and Fig.~\ref{fig:robot_platform} (c). Such embodied dynamics introduce challenges for robust trajectory modeling, accurate association, and reliable tracking performance across diverse real-world environments and complex scenarios.

EmboTrack maintains consistent bounding box size distribution across its training and test splits, ensuring balanced representation of object scales. As shown in Fig.~\ref{fig:distribution_box}, the bounding box size distributions of both QuadTrack and BipTrack exhibit strong alignment between training and test subsets. This indicates that object scale variations are evenly represented, minimizing distribution bias and ensuring reliable generalization for model evaluation. Such consistency is crucial for panoramic MOT, where object appearance and scale can fluctuate drastically due to viewpoint distortion and motion dynamics. The balanced statistical design of EmboTrack thus provides a stable foundation for fair and reproducible algorithm benchmarking.

%% file: Table_Dataset_Comparison.tex
\begin{table*}[htbp]
	\centering
	\small
	\resizebox{0.98\textwidth}{!}{
		\setlength\tabcolsep{8pt}
		\renewcommand\arraystretch{1.0}
    \begin{tabular}{l@{}lc|cc|cc|cccc}
        % \toprule
        % \hline
        \topline
        \rowcolor{poster_color1}
          &&& \multicolumn{2}{c|}{\textbf{Data}} & \multicolumn{2}{c|}{\textbf{Domain}}   & &&&\\
        \rowcolor{poster_color1}&
          \multirow{-2}{*}{ \textbf{Datasets}}   & \multirow{-2}{*}{ \textbf{Pub}} &  \textbf{Cov.}  & \textbf{Pano.}  & \textbf{Platform}  & \textbf{Movement}  &  \multirow{-2}{*}{\textbf{Trk Len}} & \multirow{-2}{*}
        {\textbf{No. Seq}} &\multirow{-2}{*}{\textbf{No. Smp}} &\multirow{-2}{*}{\textbf{No. T}} \\

        % \midrule 
        \hline\hline
        
\multicolumn{2}{l}{\textbf{KITTI MOT}~\cite{geiger2013vision}}                    & 2012  &   \textit{n.a.}   & \crossmark &  \car & \wheels &  \textit{n.a.}  &  21    &   8K       &  749       \\

     \multicolumn{2}{l}{\textbf{MOT17}   \cite{milan2016mot16} }             & 2017 &   \textit{n.a.}   & \crossmark &  \webm & \gait \stationary   & $\leq$85s  &   14   &       11K      &  1331       \\

    \multicolumn{2}{l}{\textbf{BDD100K}  \cite{bdd100k}  }               & 2018 &    \textit{n.a.}  & \crossmark &  \car  &  \wheels & 40s  &   2000   &        398K     &    \textit{n.a.}    \\
         
     \multicolumn{2}{l}{\textbf{JRDB} \cite{martin2021jrdb} }                 & 2019 &  360$^{\circ}$    & \crossmark &  \robot &  \wheels  &  $\leq$117s &   54   &      20K       &     \textit{n.a.}    \\
     
       \multicolumn{2}{l}{\textbf{nuScenes}  \cite{caesar2020nuscenes}}         & 2019 &             360$^{\circ}$    & \crossmark & \car  & \wheels &  20s &   1000   &     40K  & \textit{n.a.}    \\

  \multicolumn{2}{l}{\textbf{Waymo}  \cite{waymo}}                      & 2019 &  220$^{\circ}$ & \crossmark &  \car &  \wheels  & 20s & 103K     &      20M       &     \textit{n.a.}    \\
  
   \multicolumn{2}{l}{\textbf{MOT20}  \cite{dendorfer2020mot20}  }             & 2020 &  \textit{n.a.}    & \crossmark &  \webm & \stationary    & $\leq$133s &   8   &    13K     & 3833   \\
  
     \multicolumn{2}{l}{\textbf{DanceTrack}   \cite{peize2021dance}}               & 2022 &  \textit{n.a.}    & \crossmark &  \webm &  \stationary   & \textit{n.a.}  &  100    &      105K       & 990       \\
   
     \multicolumn{2}{l}{\textbf{SportsMOT}   \cite{cui2023sportsmot}}               & 2023 &    \textit{n.a.}  & \crossmark &  \webm & \stationary    & \textit{n.a.} &   240   &      150K       &   3401      \\

           % \midrule 

    \multicolumn{2}{l}{\textbf{BEE24}  \cite{10851814} }            & 2024 &  \textit{n.a.}    & \crossmark &  \webm & \stationary    & $\leq$3s &   36   &    23K     & 4559  \\
           \hline

\multirow{2}{*}{% 
\rotatebox[origin=c]{90}{\makecell[cc]{\textbf{Embo}\\\textbf{Track}}}% 
}
    &  \textbf{~QuadTrack (ours)} & 2025 & 360° & \mycheckmark & \robotdog & \gait & $\leq$60s & 32 & 19K & 332 \\    
 &  \textbf{~BipTrack (ours)}  & 2025 & 360° & \mycheckmark &  \wheelleg & \gait & $\leq$60s & 12 & 7K & 278 \\   

        % \bottomrule
        \bottomline
        % \hline
        
    \end{tabular}
	}
 % \vspace{-3mm}
	\captionsetup{font=footnotesize}
	\caption{Typical datasets for 2D tracking. Abbreviations:  \car~(Autonomous Car), \robot~(Mobile Robot), \robotdog~(Quadruped Robot), \wheelleg~(Wheeled-Legged Robot) \mywebm~(Internet images/videos), \wheels~(Wheels), \gait~(Gait), \stationary~(Stationary), Cov. (Coverage), Pano. (Panoramic camera), %
 Trk Len (Track Length), No. Seq (The number of sequences), No. Smp (The number of samples), and No. T (the number of tracks).}
        % \vspace{-3mm}
	\label{tab:comparison dataset}
	% \vspace{-8pt}
\end{table*}

%% file: Sec_Experiments.tex
\subsection{Setups}
\subsubsection{Datasets.}

We evaluate our approach on two panoramic multi-object tracking benchmarks: JRDB~\cite{martin2021jrdb} and our proposed EmboTrack.
JRDB is a large-scale dataset captured in crowded human-centric environments, consisting of $10$ training, $7$ validation, and $27$ test sequences. The panoramic frames are generated via multi-lens stitching from five pinhole cameras mounted on a wheeled robotic platform. 
It covers both indoor and outdoor scenes featuring frequent occlusions, dense crowds, and small-scale objects. Moreover, the dataset includes targets exhibiting rapid relative motion to the robot, posing notable challenges for robust MOT performance.
EmboTrack is a newly constructed panoramic multi-object tracking dataset captured using a single $360^{\circ}$ camera. 
It consists of two sub-datasets, \ie, QuadTrack (captured directly via a panoramic annular lens) and BipTrack (generated by stitching dual-fisheye lenses from an Insta360 camera), recorded respectively from quadrupedal and wheel-legged robotic platforms, each characterized by distinct locomotion dynamics and motion perturbations. These variations introduce diverse egocentric motion patterns and trajectory behaviors, providing a more comprehensive evaluation basis for embodied panoramic tracking. Details of each subset and the integrated benchmark are provided in Sec.~\ref{sec:QuadTrack}.

\begin{table}[h]
\centering

\label{tab:panorama_generation}
\resizebox{0.95\columnwidth}{!}{
\begin{tabular}{llcc}
\toprule
Dataset & Subset & Platform & Panorama Generation Method \\
\midrule
JRDB & - & Wheeled~\robot & Multi-lens Stitching (5 pinholes) \\
\midrule
\multirow{2}{*}{EmboTrack (Ours)} & QuadTrack & Quadrupedal~\robotdog & Panoramic Annular Lens (PAL) \\
 & BipTrack & Wheel-legged~\wheelleg & Dual-fisheye Stitching (Insta360) \\
\bottomrule
\end{tabular}
}
\caption{Overview of the panoramic MOT benchmarks and imaging platforms employed in our study. The comparison underscores variations in panoramic imaging modalities and platform-specific locomotion across the evaluated datasets.}
\end{table}

\subsubsection{Metrics.}

We adopt a comprehensive suite of established multi-object tracking metrics for balanced evaluation. Specifically, we report the CLEAR MOT measures~\cite{bernardin2008evaluating}—Multi-Object Tracking Accuracy (MOTA), Detection Accuracy (DetA), and Association Accuracy (AssA)—along with IDF1~\cite{ristani2016performance}, Optimal Sub-Pattern Assignment (OPSA)~\cite{martin2021jrdb}, and Higher Order Tracking Accuracy (HOTA)~\cite{luiten2021hota}.
MOTA provides an overall summary sensitive to detection quality, while DetA and AssA separately evaluate detection and association performance. IDF1 complements them by emphasizing identity preservation over full trajectories. OPSA extends traditional IoU-based evaluation to a set-level comparison, assessing pattern-level consistency between predicted and ground-truth detections. HOTA unifies detection, localization, and association quality into a single formulation, offering a comprehensive view of tracking performance.

\input{Table_CA_EmboTrack_test}

% \vspace{-3mm}
\subsubsection{Implementation details.}
We conduct experiments on three panoramic multi-object tracking datasets, \ie, JRDB~\cite{martin2021jrdb}, QuadTrack, and BiqTrack, and train and evaluate all models independently on their respective datasets to ensure fair, dataset-specific comparisons. Our proposed OmniTrack++ comprises three configurations: OmniTrack++$_{E2E}$, which follows an End-to-End paradigm where tracks are updated via confidence-driven predictions; OmniTrack++$_{DA}$, which follows a Tracking-By-Detection (TBD) paradigm and updates tracks through explicit data association; and OmniTrack++$_{Det}$, a detection-only variant obtained by disabling the track-management module to produce single-frame outputs. 
All models use a ResNet-50 backbone~\cite{He_2016_CVPR} and an input resolution of $1536{\times}800$, with $300$ object queries, a batch size of $1$, and both the detection-confidence and track-update thresholds set to $0.50$. Training is performed for $20$ epochs using the AdamW optimizer with an initial learning rate of $2{\times}10^{-4}$; 
for the ablation experiments reported in Tab.~\ref{tab:model_components}, \ref{tab: FlexiTrack_instance}, \ref{tab:CSEM}, \ref{tab:dssm}, models were trained for $5$ epochs.
All implementations are in PyTorch, and experiments were carried out on four NVIDIA A6000 GPUs with standardized settings to support reproducible and meaningful evaluation.

\subsection{Benchmarking and Comparative Analysis}

\subsubsection{Tracking on EmboTrack test set.}

Tab.~\ref{tab:sota QaudTrack} summarizes the tracking performance on the EmboTrack test set. On QuadTrack, OmniTrack++$_{E2E}$ attains a HOTA of $34.90$, yielding an absolute improvement of $+15.03$ points (from $19.87$ to $34.90$) compared to the original OmniTrack, while OmniTrack++$_{DA}$ further elevates the score to $36.08$. These notable gains can be attributed to the integration of the ExpertTrack Memory and the refined Tracklets Management, which effectively mitigate occlusion-induced identity fragmentation and maintain stable associations under dynamic egocentric motion. On BipTrack, which establishes a new benchmark for wheel-legged robotic tracking, OmniTrack++$_{E2E}$ and OmniTrack++$_{DA}$ achieve $44.63$ and $44.96$ HOTA, respectively. Notably, our methods demonstrate state-of-the-art performance in primary association metrics (i.e., HOTA and IDF1) across the board. The results across both sub-datasets firmly verify the superior adaptability and robustness of our panoramic MOT framework under diverse and challenging motion dynamics.
\input{Table_CA_JRDB_test}

\input{Table_CA_TBD_Mode}
\input{Table_CA_E2E_Mode}

\subsubsection{Tracking on JRDB test set.} 

In Tab.~\ref{tab:sota JRDB}, we present a comprehensive comparison of our proposed OmniTrack++ against state-of-the-art methods on the JRDB test set. Notably, OmniTrack++$_{E2E}$ demonstrates substantial improvements over the original OmniTrack$_{E2E}$, achieving a HOTA score of $25.50$ and an IDF1 of $28.00$, representing absolute improvements of $+3.94$ and $+5.13$ points, respectively. 
These results highlight the effectiveness of our enhancements with the E2E tracking paradigm and bring the performance closer to that typically observed under the TBD framework. Under the TBD paradigm, while OmniTrack++$_{DA}$ experiences a slight drop in IDF1 and MOTA compared to OmniTrack$_{DA}$, it still achieves state-of-the-art HOTA ($27.03$) and OSPA ($0.81$) on the JRDB dataset. This minor trade-off in specific metrics is expected, as the design of OmniTrack++ primarily focuses on optimizing E2E tracking; consequently, certain aspects of data-association integration receive less targeted refinement. Overall, the results confirm that OmniTrack++ consistently advances tracking accuracy, particularly in the E2E setting, while maintaining competitive performance in TBD evaluation.

\subsubsection{Tracking on OmniTrack TBD with Feedback.}

On the JRDB~\cite{martin2021jrdb} validation set, we conduct a detailed analysis of our OmniTrack++ algorithm under the TBD paradigm, as summarized in Tab.~\ref{tab:tbd_mode}. Here, the baseline uses the YOLO26~\cite{yolo26} detector combined with representative trackers, ByteTrack~\cite{zhang2022bytetrack} and HybridSORT~\cite{yang2024hybrid}. Comparing the first and third rows, it is evident that our OmniTrack++$_{Det}$ significantly outperforms YOLO26, yielding notable improvements in both HOTA and IDF1. Furthermore, the comparison between the third and fifth rows shows that incorporating the feedback mechanism consistently improves tracking performance: within the OmniTrack++ framework, HOTA and IDF1 on ByteTrack increase by $+0.48$ and $+0.48$ points, respectively, while on HybridSORT, HOTA improves by $+0.28$ points and IDF1 by $+0.68$ points. These results indicate that the feedback mechanism, by leveraging instance-level information from previous frames, contributes to measurable gains in tracking metrics under the TBD paradigm.

\subsubsection{Tracking in OmniTrack E2E Mode.}

As shown in Tab.~\ref{tab:E2E_mode}, we further evaluate the performance of our OmniTrack++ algorithm under the E2E paradigm on the JRDB validation set. Compared to the original OmniTrack, the model size increases from $63.13\text{M}$ to $70.05\text{M}$ parameters, representing an approximate $11\%$ increase. Despite this moderate growth in model complexity, OmniTrack++$_{E2E}$ achieves a remarkable improvement in tracking performance, with HOTA and IDF1 yielding absolute improvements of $+5.72$ and $+8.24$ points, respectively. Notably, these gains not only demonstrate the effectiveness of the proposed enhancements but also surpass the current state-of-the-art E2E tracker, MeMOTR, highlighting the substantial advantage of OmniTrack++ in E2E MOT on JRDB. This significant improvement is largely attributed to our ExpertTrack Memory, which provides high-quality trajectory instance information. By effectively capturing and leveraging these exemplar trajectories, OmniTrack++ can more accurately associate and maintain object tracks over time, thereby contributing directly to the observed gains in tracking performance.

\input{Table_AS_Components}

\subsection{Ablation Studies}
\subsubsection{Ablation Analysis of Model Components}

To quantify the contribution of each proposed component, we conduct a comprehensive ablation study on the JRDB validation set after $5$ training epochs by progressively removing DynamicSSM (DSSM; Sec.~\ref{subsec:CSEM}) and ExpertTrack Memory (ETM; Sec.~\ref{subsec:ExpertTrack}) from the full framework, as summarized in Tab.~\ref{tab:model_components}. The results show that incorporating ETM alone leads to a modest improvement of $+0.31$ points in HOTA and $+0.02$ points in IDF1, while DSSM alone yields more substantial gains of $+1.04$ points in HOTA and $+1.10$ points in IDF1, demonstrating the effectiveness of each component in enhancing primary tracking performance. When both components are combined, the improvements increase further to $+1.17$ points in HOTA and $+1.49$ points in IDF1, confirming that DSSM and ETM complement each other and jointly contribute to more robust and accurate temporal associations. It is worth noting that while the introduction of ETM leads to a decrease in the MOTA score (from $27.34$ to $22.20$ in the full model)—a common trade-off since memory mechanisms can be more sensitive to false positives—our framework achieves optimal performance across the more comprehensive association metrics (HOTA, IDF1, and OSPA), which better reflect the true tracking quality in complex panoramic scenes.

\input{Table_AS_Temporal_Query}

\input{Table_AS_Comparison_of_DSSM_Block}

\subsubsection{Analysis of the FlexiTrack Instance.}

To evaluate the impact of FlexiTrack instances on E2E MOT performance during the training phase, we design an ablation study comparing FlexiTrack instances (\(\textit{I}_{ft}\)) and denoised instances (\(\textit{I}_{dn}\)), with results summarized in Tab.~\ref{tab: FlexiTrack_instance}. 
Here, \(\textit{I}_{ft}\) refers to instances generated via our feedback mechanism (Sec.~\ref{subsec:Framework}), while \(\textit{I}_{dn}\) are derived from the Ground Truth (GT) by introducing random perturbations. 
In Exp. \tikz[baseline, yshift=0.6ex]{\node[draw,circle,inner sep=1.1pt, font=\tiny] {1};}, where both \(\textit{I}_{dn}\) and \(\textit{I}_{ft}\) are absent during training, tracking fails entirely, as expected, since the network lacks any instance-level association cues. 
In Exp. \tikz[baseline, yshift=0.6ex]{\node[draw,circle,inner sep=1.1pt, font=\tiny] {2};}, using only \(\textit{I}_{ft}\) during training yields a modest HOTA of $4.39$, indicating that while feedback provides some informative signals, establishing accurate associations remains challenging and prone to overfitting. 
In contrast, Exp. \tikz[baseline, yshift=0.6ex]{\node[draw,circle,inner sep=1.1pt, font=\tiny] {3};}, which utilizes only \(\textit{I}_{dn}\) in training, achieves a substantial improvement to $14.64$ in HOTA, reflecting the strong association cues inherited from GT-based information that effectively guide the network in linking simple targets. 
Finally, incorporating both \(\textit{I}_{dn}\) and \(\textit{I}_{ft}\) during training in Exp. \tikz[baseline, yshift=0.6ex]{\node[draw,circle,inner sep=1.1pt, font=\tiny] {4};} further boosts performance to $28.47$ in HOTA, demonstrating that the addition of \(\textit{I}_{ft}\) significantly enhances the accuracy of instance association by complementing the denoised GT signals. 
Overall, these results highlight the synergistic effect of combining denoised and feedback-generated instances during training in improving E2E tracking performance.

\input{Table_AS_DynamicSSM}
\input{Table_AS_Epoch}

\subsubsection{Analysis of the DynamicSSM Block.}

In Tab.~\ref{tab:CSEM}, we conduct an ablation study to evaluate the effectiveness of the DynamicSSM Block across different feature levels ($S_5$, $S_4$, and $S_3$), compared with conventional convolutional (Conv) and multilayer perceptron (MLP) layers.
Experiments \tikz[baseline, yshift=0.6ex]{\node[draw,circle,inner sep=1.1pt, font=\tiny]{1};}--\tikz[baseline, yshift=0.6ex]{\node[draw,circle,inner sep=1.1pt, font=\tiny]{3};} show that replacing all features with Conv or MLP layers leads to suboptimal results (HOTA $23.19$–$27.61$, IDF1 $26.51$–$31.21$), revealing the limitations of these operators in modeling dynamic spatio-temporal dependencies.
In contrast, selectively introducing DynamicSSM yields consistent improvements: applying it to $S_5$ and $S_3$ individually (\tikz[baseline, yshift=0.6ex]{\node[draw,circle,inner sep=1.1pt, font=\tiny]{5};}, \tikz[baseline, yshift=0.6ex]{\node[draw,circle,inner sep=1.1pt, font=\tiny]{7};}) improves both HOTA and IDF1, while its use at $S_4$ (\tikz[baseline, yshift=0.6ex]{\node[draw,circle,inner sep=1.1pt, font=\tiny]{6};}) achieves the best performance ($28.47$ HOTA, $32.68$ IDF1, OSPA $0.8552$).
This suggests that $S_4$, integrating high-level semantics with mid-level geometry, benefits most from DynamicSSM’s temporal modeling capability, thereby enhancing instance association and overall tracking performance.

\begin{figure}[!t]
    % \vskip -2 ex
    \centering
    \includegraphics[width=\linewidth]{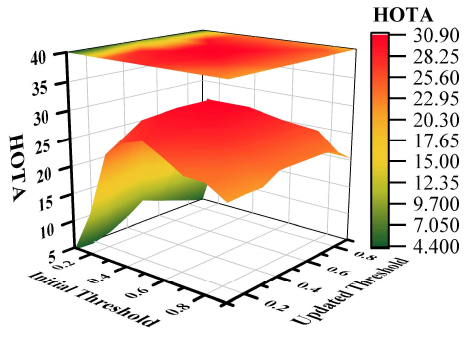}
    \caption{Effects of the trajectory initialization threshold and update threshold on the HOTA metric in OmniTrack++$_{E2E}$.}
    \label{fig: thresholds}
    % \vskip -2 ex
\end{figure} 

In Tab.~\ref{tab:dssm}, we further analyze the contributions of individual components within the DynamicSSM Block to the overall performance. As shown, the baseline model without Dconv, SSM, or Fusion (Exp.\tikz[baseline, yshift=0.6ex]{\node[draw,circle,inner sep=1.1pt, font=\tiny] {1};}) achieves a HOTA of $27.61$, IDF1 of $31.21$, and OSPA of $0.8610$. Incorporating only the SSM and Fusion modules (Exp.\tikz[baseline, yshift=0.6ex]{\node[draw,circle,inner sep=1.1pt, font=\tiny] {2};}) results in a minor decrease in HOTA, suggesting that Dconv plays a complementary role in enhancing performance. When using Dconv together with either SSM (Exp.\tikz[baseline, yshift=0.6ex]{\node[draw,circle,inner sep=1.1pt, font=\tiny] {4};}) or Fusion (Exp.\tikz[baseline, yshift=0.6ex]{\node[draw,circle,inner sep=1.1pt, font=\tiny] {3};}), the results highlight that each component contributes uniquely to the model's effectiveness, though full integration of all three modules (Exp.\tikz[baseline, yshift=0.6ex]{\node[draw,circle,inner sep=1.1pt, font=\tiny] {5};}) yields the best performance with HOTA of $28.47$, IDF1 of $32.68$, and OSPA of $0.8552$. These observations indicate that Dconv, SSM, and Fusion synergize to improve tracking accuracy, with the complete DynamicSSM Block consistently outperforming all partial configurations, thus validating the design choice of integrating all components.

% \vspace{-3mm}
\subsubsection{Analysis of the initialization and update thresholds.}
Fig.~\ref{fig: thresholds} illustrates the influence of the trajectory initialization and update thresholds on the HOTA metric in OmniTrack++$_{E2E}$. We observe that HOTA is highly sensitive to the initialization threshold, particularly in the range of $0.3{\sim}0.6$, where both under- and over-initialization can significantly degrade performance. A low initialization threshold tends to introduce noisy or short-lived tracks, reducing identity consistency, whereas an excessively high threshold delays track activation and causes missed detections during early motion stages. Similarly, the update threshold relates non-linearly to performance: overly strict updates suppress track continuity, while overly loose updates increase identity switches. The optimal region, corresponding to balanced precision and continuity, is observed around an initialization threshold of about $0.5$ and an update threshold around $0.5$, validating the effectiveness of our confidence-based update mechanism in stabilizing trajectory management.
\begin{figure}[!t]
    % \vskip -2 ex
    \centering
    \includegraphics[width=\linewidth]{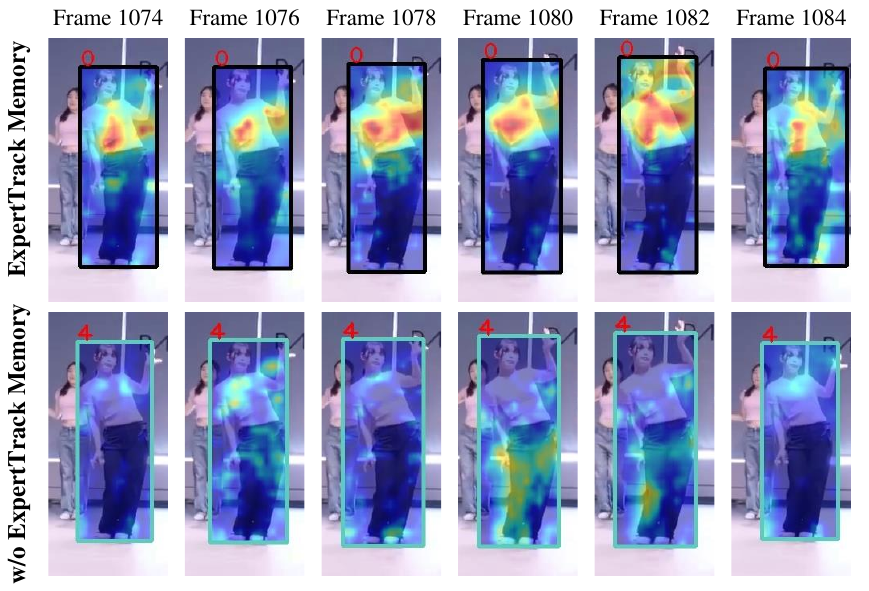}
    \caption{Visualization of query localization. The first row shows heatmaps with the proposed \emph{ExpertTrack Memory}, where trajectory-informed feedback maintains consistent focus on target regions. The second row shows results without it, where localization is unstable and dispersed across frames.}
    \label{fig:cam}
    % \vskip -2 ex
\end{figure} 

\begin{figure*}[htbp]
  \centering
  \includegraphics[width=0.98\textwidth]{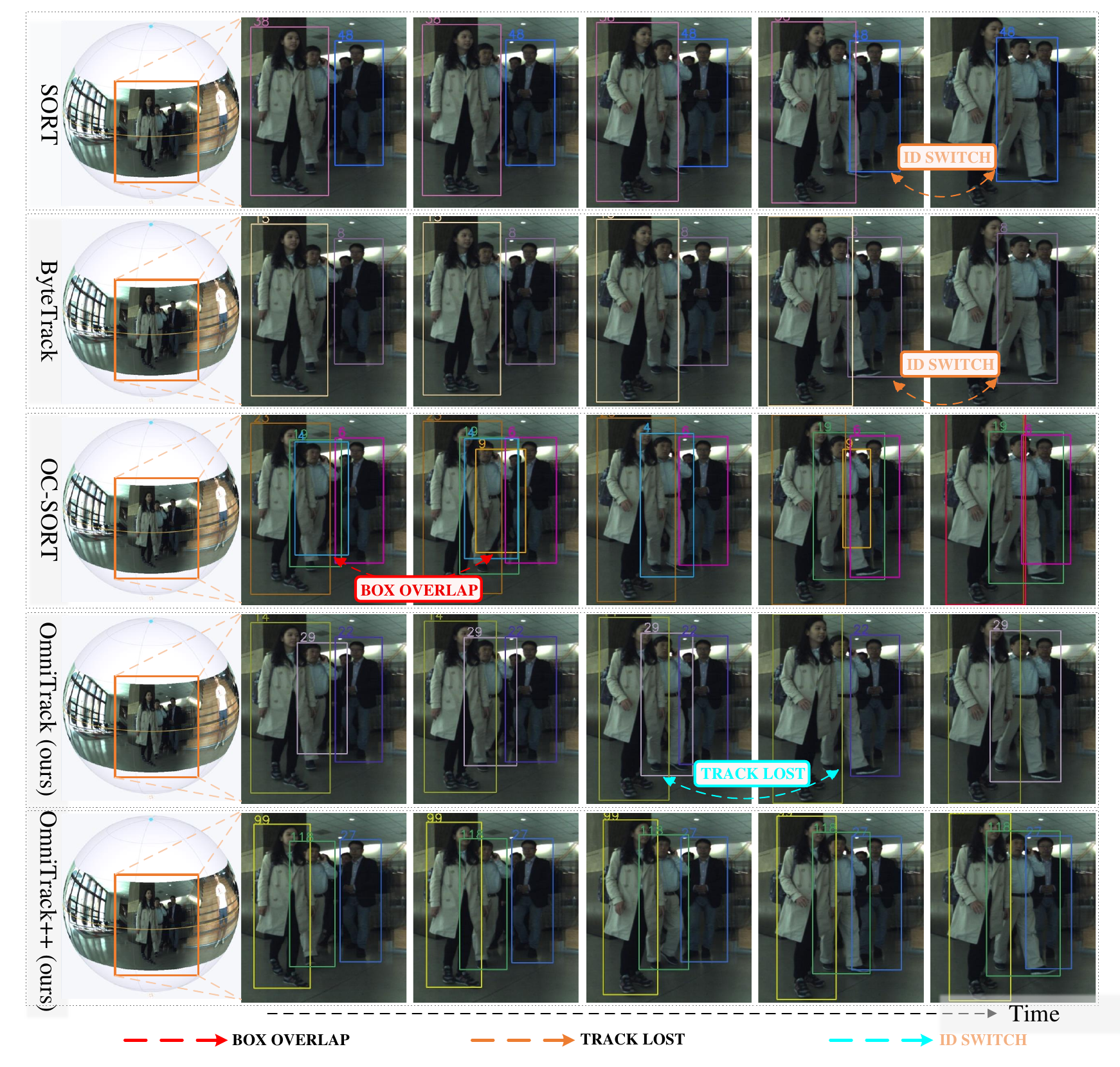}
  \vskip 2ex
  \caption{Comparison of different MOT methods \cite{wojke2017simple,zhang2022bytetrack,cao2023observation} on the JRDB dataset \cite{martin2021jrdb}, visualized for frames 300$\sim$310 (every other frame) of the sequence \emph{nvidia-aud-2019-04-18\_0}. 
  As shown in the visualizations, OmniTrack$++$ demonstrates robust tracking performance, effectively maintaining consistent associations even under challenging conditions such as occlusions and motion dynamics.}
  \label{fig: vis tracking}
  \vskip 4ex
\end{figure*}

\begin{figure*}[htbp]
  \centering
  \includegraphics[width=0.98\textwidth]{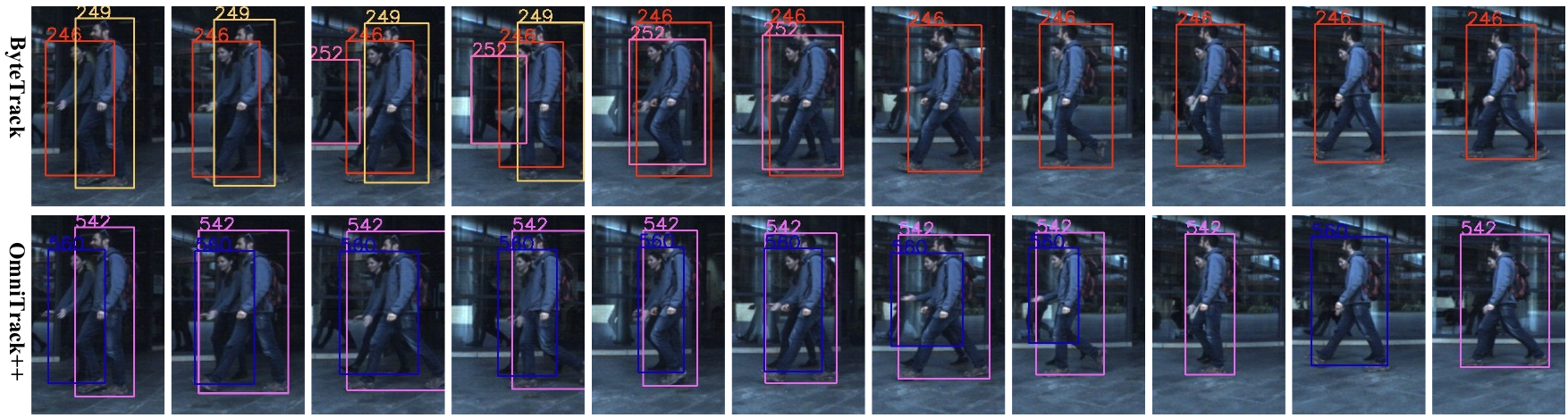}
  \vskip 3ex
  \caption{Analysis of failure cases on JRDB~\cite{martin2021jrdb}, highlighting scenarios where OmniTrack++ struggles compared to ByteTrack~\cite{zhang2022bytetrack}.}
  \label{fig:failure}
  \vskip 4ex
\end{figure*}

\subsubsection{Analysis of Performance Across Different Epochs.}

We conducted an analysis of our model across different training epochs on the JRDB validation set, with the results summarized in Table~\ref{tab:epoch}. 
As shown, the model exhibits a notable performance jump at epoch~$5$, where HOTA increases from $24.20$ to $28.47$, yielding an improvement of $4.27$. A slight drop occurs at epoch~$6$, followed by a gradual improvement, reaching the peak performance at epoch~$16$ with HOTA of $30.84$ and IDF1 of $35.66$. Subsequent training leads to a slow decline, stabilizing around $29$ HOTA by epoch~$20$, which indicates that the model achieves a stable convergence trend, demonstrating both rapid initial learning and gradual stabilization over the course of training.

\subsection{Visualization Analysis}

The Fig.~\ref{fig:cam} demonstrates the effectiveness of \emph{ExpertTrack Memory} in stabilizing query localization. In the first row, the heatmaps exhibit highly concentrated and intense activations on the target's core, maintaining a stable focal point despite significant pose variations from Frame 1074 to 1084. In contrast, without this module (second row), the energy distributions become dispersed and fluctuate across frames, often drifting toward background regions. This comparison highlights that the trajectory-informed feedback provides essential structural guidance, ensuring consistent identity preservation and robust localization in complex scenarios.

The Fig.~\ref{fig: vis tracking} provides a qualitative comparison of different MOT algorithms on the JRDB dataset~\cite{martin2021jrdb}. As shown in the first row, SORT~\cite{bewley2016simple} fails to maintain the trajectory of a partially occluded target and further suffers from ID switches in the last two frames. A similar issue can be observed in the second row with ByteTrack~\cite{zhang2022bytetrack}, where partial occlusion leads to trajectory fragmentation and identity inconsistency. In contrast, OC-SORT~\cite{cao2023observation} (third row) successfully preserves the target trajectory under occlusion, yet introduces overlapping bounding boxes and false positives. The fourth row presents the results of our proposed OmniTrack~\cite{luo2025omnidirectional}, which achieves more stable tracking overall, with only a brief trajectory gap in the middle frames due to partial occlusion. Finally, the last row demonstrates OmniTrack++, which effectively addresses the trajectory loss observed in OmniTrack. Benefiting from the ExpertTrack Memory module, OmniTrack++ produces a robust FlexiTrack Instance, enabling accurate recall of occluded targets and maintaining consistent trajectories across challenging scenarios.

\subsection{Failure Case Analysis}
To further examine the limitations of our approach, we analyze several representative failure cases on the JRDB dataset. 
As illustrated in Fig.~\ref{fig:failure}, our method occasionally struggles in complex real-world scenarios involving heavy occlusion, dense crowds, or abrupt camera motion. Specifically, our tracker may produce fragmented trajectories or temporary identity switches when multiple pedestrians overlap or move in close proximity. 
These issues mainly arise from severe occlusion and motion ambiguity, which can disrupt stable target association. Nevertheless, compared with ByteTrack, our method exhibits better robustness in maintaining identity consistency and recovering from short-term tracking interruptions, demonstrating the effectiveness of our dynamic association design. These observations highlight the remaining challenges in crowded or highly dynamic scenes and provide valuable insights for future improvements, such as introducing more explicit temporal reasoning and occlusion-aware modeling to further enhance tracking stability.

%% file: Table_CA_EmboTrack_test.tex
% \begin{table}[htbp]
\begin{table}[t]
    \centering
    \setlength{\tabcolsep}{4pt}
    \resizebox{\columnwidth}{!}{%
    \begin{tabular}{c|l|cccc}
        % \toprule
        \topline

        \rowcolor{mygray} &  Method  & {HOTA}$\uparrow$ & {OSPA}$\downarrow$ & {IDF1}$\uparrow$ & {MOTA} $\uparrow$  \\
        % \midrule 
        \hline
        
        \multicolumn{6}{c}{\textbf{QuadTrack Dataset}}  \\
        \hline
        
       \parbox[t]{3mm}{\multirow{4}{*}{\rotatebox[origin=c]{90}{E2E}}} & TrackFormer~\cite{meinhardt2021trackformer} &  19.62 & 0.97  & 17.75 &  3.16  \\
        & MOTRv2~\cite{zhang2023motrv2}    & 16.42 & 0.96   & 17.08  & -0.06\\
        & \cellcolor{tabgray}\name{}$_{E2E}$ \text{(ours)} & \cellcolor{tabgray}19.87 & \cellcolor{tabgray}0.98 & \cellcolor{tabgray}19.47 & \cellcolor{tabgray}-5.89  \\

        & \cellcolor{tabgray}\name{}++$_{E2E}$ \text{(ours)} & \cellcolor{tabgray}\textbf{34.90} & \cellcolor{tabgray}\textbf{0.85} & \cellcolor{tabgray}\textbf{41.21} & \cellcolor{tabgray}\textbf{18.65}  \\

        % \midrule  	 	 	 

        \hline
        \parbox[t]{3mm}{\multirow{9}{*}{\rotatebox[origin=c]{90}{TBD}}} 
        & SORT~\cite{bewley2016simple}    & 14.57  & 0.98  & 15.60 & 4.81  \\
        & DeepSORT~\cite{wojke2017simple} & 21.16  & 0.96  & 22.56 & 5.12  \\
        & ByteTrack~\cite{zhang2022bytetrack}   & 20.66  & 0.94  & 22.56 & 8.68  \\
        & Bot-SORT~\cite{aharon2022bot} & 15.77  & 0.99  & 15.65 & 5.92  \\
        & OC-SORT~\cite{cao2023observation}&  20.83 & 0.94 & 22.60 & 7.65  \\ 	 	 	

        & HybridSORT~\cite{yang2024hybrid}  &  16.64  & 0.96  & 17.38 & 6.79  \\
        & DiffMOT~\cite{lv2024diffmot} &  16.40 & 0.97  & 16.62 & 6.21 \\ 	 	 	 

        & \cellcolor{tabgray}\name{}$_{DA}$ \text{(ours)} & \cellcolor{tabgray}23.45 & \cellcolor{tabgray}0.94 & \cellcolor{tabgray}26.41 & \cellcolor{tabgray}9.68 \\

        & \cellcolor{tabgray}\name{}++$_{DA}$ \text{(ours)} & \cellcolor{tabgray}\textbf{36.08} & \cellcolor{tabgray}\textbf{0.82} & \cellcolor{tabgray}\textbf{42.76} & \cellcolor{tabgray}\textbf{21.94} \\

        % \bottomrule
        \hline

                \multicolumn{6}{c}{\textbf{BipTrack Dataset}}  \\
        \hline
        
       \parbox[t]{3mm}{\multirow{4}{*}{\rotatebox[origin=c]{90}{E2E}}} &           MOTRv2~\cite{zhang2023motrv2}    & 39.29 & 0.78 & 38.60 & 2.72  \\
        & MeMOTR~\cite{MeMOTR}    &  43.17 & \textbf{0.82} & 46.22 & \textbf{27.85}   \\
        & \cellcolor{tabgray}\name{}$_{E2E}$ \text{(ours)}  &  \cellcolor{tabgray}35.70 & \cellcolor{tabgray}0.89  & \cellcolor{tabgray}33.91 &  \cellcolor{tabgray} -16.30 \\

        & \cellcolor{tabgray}\name{}++$_{E2E}$ \text{(ours)} & \cellcolor{tabgray}\textbf{44.63}&\cellcolor{tabgray}0.84&\cellcolor{tabgray}\textbf{46.81}&\cellcolor{tabgray}21.63 \\

        % \midrule  	 	 	 

        \hline
        \parbox[t]{3mm}{\multirow{9}{*}{\rotatebox[origin=c]{90}{TBD}}} 
        & SORT~\cite{bewley2016simple}    & 42.67 &0.86 &44.96 &\textbf{28.27}   \\
        & DeepSORT~\cite{wojke2017simple} & 41.15 &0.90 &38.56 &22.61 \\
        & ByteTrack~\cite{zhang2022bytetrack}   & 44.10 &0.84 &46.25&20.61 \\
        & Bot-SORT~\cite{aharon2022bot} & 42.46 &0.86 &40.88&25.72  \\
        & OC-SORT~\cite{cao2023observation}&  40.93 &0.87 &40.34 &0.45 \\ 	 
        & HybridSORT~\cite{yang2024hybrid} &  42.76 &0.85 &43.19 &13.02   \\
        & DiffMOT~\cite{lv2024diffmot}& 39.28 &0.95 &34.33 &24.54   \\	 	 	 

        & \cellcolor{tabgray}\name{}$_{DA}$ \text{(ours)} &  \cellcolor{tabgray}32.85 & \cellcolor{tabgray}0.94  & \cellcolor{tabgray}30.37 &  \cellcolor{tabgray}-32.40  \\

        & \cellcolor{tabgray}\name{}++$_{DA}$ \text{(ours)} & \cellcolor{tabgray}\textbf{44.96} &\cellcolor{tabgray}\textbf{0.76}&\cellcolor{tabgray}\textbf{47.41}&\cellcolor{tabgray}21.50  \\

        % \bottomrule
        \bottomline
        
    \end{tabular}
    } 	 	 	 
	 	 	 
    % \vspace{-1ex}
    \caption{Comparison with state-of-the-art methods on the EmboTrack test set.}
    % \vspace{-3mm}
    \label{tab:sota QaudTrack}
\end{table}

%% file: Table_CA_JRDB_test.tex
% \begin{table}[htbp]
\begin{table}[t]
    \centering
    \setlength{\tabcolsep}{4pt}
    \resizebox{\columnwidth}{!}{%
    \begin{tabular}{c|l|cccc}
        % \toprule
        \topline

       \rowcolor{mygray} 
       
       &  Method   & {HOTA}$\uparrow$ & {OSPA}$\downarrow$ & {IDF1}$\uparrow$ & {MOTA} $\uparrow$  \\
        % \midrule 
        \hline

       \parbox[t]{3mm}{\multirow{5}{*}{\rotatebox[origin=c]{90}{E2E}}} & TrackFormer~\cite{meinhardt2021trackformer} &  19.16 &  0.95  & 19.66 & 17.79  \\
        & MOTRv2~\cite{zhang2023motrv2}   &  18.22 & 0.93  & 19.30 & 12.30  \\
        & MeMOTR~\cite{MeMOTR}   &  25.10 & \textbf{0.87}  & 27.46 & 22.53  \\
        & \cellcolor{tabgray}\name{}$_{E2E}$ \text{(ours)} & \cellcolor{tabgray}21.56 & \cellcolor{tabgray}0.94 & \cellcolor{tabgray}22.87 & \cellcolor{tabgray}\textbf{25.01} \\

        & \cellcolor{tabgray}OmniTrack++$_{E2E}$ \text{(ours)} & \cellcolor{tabgray}\textbf{25.50} & \cellcolor{tabgray}0.88 & \cellcolor{tabgray}\textbf{28.00} & \cellcolor{tabgray}21.02 \\

        % \midrule
        \hline
        \parbox[t]{3mm}{\multirow{9}{*}{\rotatebox[origin=c]{90}{TBD}}} 
        & SORT~\cite{bewley2016simple}    &  23.49 & 0.90  & 26.11 & 24.59  \\
        & DeepSORT~\cite{wojke2017simple}  &  22.15 &  0.95 & 23.46 & 24.88 \\
        & ByteTrack~\cite{zhang2022bytetrack}   &   25.00 &  0.86  & 27.95 & 26.59 \\
        & Bot-SORT~\cite{aharon2022bot} &    22.90  & 0.91  & 24.27 & 23.08  \\
        & OC-SORT~\cite{cao2023observation} &     25.04 & 0.84  & 27.89 & 25.64  \\
        & HybridSORT~\cite{yang2024hybrid}  &  25.01  &  0.85 & 27.82  & 25.03  \\
        & DiffMOT~\cite{lv2024diffmot} &    19.96  & 0.95  & 20.26 &  20.05  \\
        & \cellcolor{tabgray}\name{}$_{DA}$ \text{(ours)} & \cellcolor{tabgray}26.92 & \cellcolor{tabgray}0.84 & \cellcolor{tabgray}\textbf{30.26} & \cellcolor{tabgray}\textbf{26.60}  \\
        & \cellcolor{tabgray}\name{}++$_{DA}$ \text{(ours)} & \cellcolor{tabgray}\textbf{27.03}  & \cellcolor{tabgray}\textbf{0.81} & \cellcolor{tabgray}29.52 & \cellcolor{tabgray}25.05  \\
        % \bottomrule
        \bottomline
    \end{tabular}
    }
    \vspace{-1ex}
    \caption{Comparison with state-of-the-art methods on the JRDB test set~\cite{martin2021jrdb}.}
    % \vspace{-2ex}
    \label{tab:sota JRDB}
\end{table}

%% file: Table_CA_TBD_Mode.tex
\begin{table*}[htbp]
    \centering
    \setlength{\tabcolsep}{4pt}
    \resizebox{2.0\columnwidth}{!}{%
    \begin{tabular}{l|cl|l|lll|lll}
        \topline
          \rowcolor{mygray}  Method & Feedback & Detector  & Tracker & HOTA $\uparrow$ & IDF1$\uparrow$ & OSPA $\downarrow$  & MOTA$\uparrow$ & DetA $\uparrow$ & FPS $\uparrow$ \\
          \hline
          
        baseline & \crossmark & YOLO26 & \multirow{5}{*}{\makecell{ByteTrack\\ \cite{zhang2022bytetrack}}} & 27.94 & 32.17 & 0.868 & 33.91 & 29.45  & 55.28\\
        
        vanilla TBD & \crossmark &   OmniTrack$_{Det}$ & & 28.14 \textcolor{codegreen}{(+0.20)} & 32.97 \textcolor{codegreen}{(+0.80)} & 0.870 \textcolor{black}{(+0.002)} & 37.36 \textcolor{codegreen}{(+3.45)} & 32.94 \textcolor{codegreen}{(+3.49)}  & 12.24\\

        vanilla TBD & \crossmark &   OmniTrack++$_{Det}$ & & 29.23 \textcolor{codegreen}{(+1.29)} & 33.74 \textcolor{codegreen}{(+1.57)} & 0.873 \textcolor{black}{(+0.005)} & 36.30 \textcolor{codegreen}{(+2.39)} & 33.68 \textcolor{codegreen}{(+4.23)}  & 16.62 \\
          
        OmniTrack$_{DA}$~(ours) & \mycheckmark &   OmniTrack$_{Det}$ & & 29.58 \textcolor{codegreen}{(+1.64)} & 34.54 \textcolor{codegreen}{(+2.37)} & 0.859 \textcolor{codegreen}{(-0.009)} & 38.14 \textcolor{codegreen}{(+4.23)} & 34.71 \textcolor{codegreen}{(+5.26)}  & 11.83\\

        OmniTrack++$_{DA}$~(ours) & \mycheckmark &   OmniTrack++$_{Det}$ & & 29.71 \textcolor{codegreen}{(+1.77)} & 34.22 \textcolor{codegreen}{(+2.05)} & 0.873 \textcolor{black}{(+0.005)} & 30.82 \textcolor{black}{(-3.09)} & 31.89 \textcolor{codegreen}{(+2.44)}  & 13.54\\

        \hline

        baseline & \crossmark & YOLO26 & \multirow{5}{*}{\makecell{HybridSORT\\ \cite{yang2024hybrid}}} & 29.93 & 34.18 & 0.860 & 34.61 & 31.16  & 50.74\\
        
        vanilla TBD & \crossmark &   OmniTrack$_{Det}$ & & 30.00 \textcolor{codegreen}{(+0.07)} & 34.09 \textcolor{black}{(-0.09)} & 0.853 \textcolor{codegreen}{(-0.007)} & 32.32 \textcolor{black}{(-2.29)} & 35.02 \textcolor{codegreen}{(+3.86)}  & 11.65\\

        vanilla TBD & \crossmark &   OmniTrack++$_{Det}$ & & 30.84 \textcolor{codegreen}{(+0.91)} & 35.50 \textcolor{codegreen}{(+1.32)} & 0.875 \textcolor{black}{(+0.015)} & 35.94 \textcolor{codegreen}{(+1.33)} & 35.55 \textcolor{codegreen}{(+4.39)}  & 15.37\\
         
        OmniTrack$_{DA}$~(ours) & \mycheckmark &   OmniTrack$_{Det}$ & & 31.05 \textcolor{codegreen}{(+1.12)} & 36.06 \textcolor{codegreen}{(+1.88)} & 0.850 \textcolor{codegreen}{(-0.010)} & 38.13 \textcolor{codegreen}{(+3.52)} & 35.08 \textcolor{codegreen}{(+3.92)}  & 10.96\\

        OmniTrack++$_{DA}$~(ours) & \mycheckmark &   OmniTrack++$_{Det}$ & & 31.12 \textcolor{codegreen}{(+1.19)} & 36.18 \textcolor{codegreen}{(+2.00)} & 0.872 \textcolor{black}{(+0.012)} & 35.30 \textcolor{codegreen}{(+0.69)} & 34.72 \textcolor{codegreen}{(+3.56)}  & 12.77\\

        \bottomline
    \end{tabular}
    }        
        
    \caption{Results on the JRDB validation set~\cite{martin2021jrdb}. The table compares four configurations: Baseline, Vanilla TBD, OmniTrack$_{DA}$, and OmniTrack++$_{DA}$.
    The Baseline employs YOLO26 as the detector with a standard Tracking-By-Detection (TBD) pipeline.
    Vanilla TBD replaces the detector with our OmniTrack$_{Det}$ or OmniTrack++$_{Det}$, enabling panoramic-aware detection while keeping the same TBD tracker. OmniTrack++$_{DA}$ builds upon OmniTrack++$_{Det}$ by incorporating our proposed feedback mechanism, where tracking outputs are fed back to the detector to refine future predictions. The \textcolor{codegreen}{numbers} represent the improvement relative to the baseline method. The FPS metric is measured on a single RTX 3090 GPU with an image resolution of $4160{\times}480$.
    }
    \label{tab:tbd_mode}
\end{table*}

%% file: Table_CA_E2E_Mode.tex
\begin{table}[htbp]
    \centering
    \setlength{\tabcolsep}{4pt}
    \resizebox{1\columnwidth}{!}{%
    \begin{tabular}{l|cccc}
        % \toprule
        \topline
          \rowcolor{mygray}  Method & Parmas & HOTA $ \uparrow$ & IDF1$\uparrow$ & OSPA $\downarrow$    \\
         % \midrule
         \hline

          TrackFormer~\cite{meinhardt2021trackformer}   & 44.01M & 22.22 & 23.38 & 0.959  \\
          MOTR~\cite{zeng2022motr}  & 43.91M   & 19.78 & 23.25 & 0.928   \\
          MOTRv2~\cite{zhang2023motrv2}  & 41.65M & 24.68 & 25.49 & {0.911}  \\
          MeMOTR~\cite{MeMOTR}   & 50.36M & 29.51 & 33.64 & 0.891    \\

        \rowcolor{tabgray}\name{}$_{E2E}$~\text{(ours)}   & 63.13M & {25.12} & {27.42} & 0.925  \\

        \rowcolor{tabgray}\name{}++$_{E2E}$~\text{(ours)}   & 70.05M & \textbf{30.84}     & \textbf{35.66}     & \textbf{0.879}  \\    

         % \bottomrule
         \bottomline
    \end{tabular}
    }        
        
    \caption{Results on the JRDB validation set~\cite{martin2021jrdb}. Comparison between the proposed OmniTrack and OmniTrack++ (E2E) and representative End-to-End multi-object tracking methods. 
    }
    % \vspace{-3mm}
    \label{tab:E2E_mode}
\end{table}

%% file: Table_AS_Components.tex
% \begin{table}[htbp]
\begin{table}[t]
    \centering
    \setlength{\tabcolsep}{8pt}     % 4pt
    \resizebox{\columnwidth}{!}{%
    \begin{tabular}{c|cc|cccc}
        % \toprule
        \topline
        \rowcolor{mygray}  Exp. & \textit{DSSM} & \textit{ETM} & HOTA$\uparrow$ & IDF1$\uparrow$ & OSPA$\downarrow$ & MOTA$\uparrow$ \\
        % \midrule
        \hline

        \tikz[baseline, yshift=0.6ex]{\node[draw,circle,inner sep=1.1pt, font=\tiny] {1};} & - & - &  27.30 & 31.19 & 0.8958  & 27.34 \\ 	 	 	 

        \tikz[baseline, yshift=0.6ex]{\node[draw,circle,inner sep=1.1pt, font=\tiny] {2};} &  & \checkmark   & 27.61 & 31.21 & 0.8610 & 11.99 \\
        \tikz[baseline, yshift=0.6ex]{\node[draw,circle,inner sep=1.1pt, font=\tiny] {3};} &   \checkmark  & & 28.34 & 32.29 & 0.8786 & \textbf{29.86} 
        \\
        % \midrule
        \hline
        \rowcolor{tabgray}\tikz[baseline, yshift=0.6ex]{\node[draw,circle,inner sep=1.1pt, font=\tiny] {4};} & \checkmark & \checkmark  & \textbf{28.47} & \textbf{32.68} & \textbf{0.8552} & 22.20 \\
        % \bottomrule
        \bottomline
    \end{tabular}
    }
    \vspace{-1mm}
    \caption{Analysis of Model Components: \(\textit{DSSM}\) represents DynamicSSM, and \(\textit{ETM}\) refers to an ExpertTrack Memory.}
    % \vspace{-3mm}
    \label{tab:model_components}
\end{table}

%% file: Table_AS_Temporal_Query.tex
% \begin{table}[htbp]
\begin{table}[t]
    \centering
    \setlength{\tabcolsep}{8pt}     % 4pt
    \resizebox{\columnwidth}{!}{%
    \begin{tabular}{c|cc|cccc}
        % \toprule
        \topline
        \rowcolor{mygray}  Exp. & \textit{I$_{dn}$} & \textit{I$_{ft}$} & HOTA$\uparrow$ & IDF1$\uparrow$ & OSPA$\downarrow$ & MOTA$\uparrow$ \\
        % \midrule
        \hline
        
        \tikz[baseline, yshift=0.6ex]{\node[draw,circle,inner sep=1.1pt, font=\tiny] {1};} & - & - &  0.01 & 0.00 & 1.0000 & 0.00  \\ 	 	 	 

        \tikz[baseline, yshift=0.6ex]{\node[draw,circle,inner sep=1.1pt, font=\tiny] {2};} & & \checkmark   & 4.39 & 1.61 & 0.9958 & -1112.10 \\

        \tikz[baseline, yshift=0.6ex]{\node[draw,circle,inner sep=1.1pt, font=\tiny] {3};} &   \checkmark  & & 14.64 & 15.11 & 0.9313  & 4.24     \\ 	 	 	

        % \midrule
        \hline
        \rowcolor{tabgray}\tikz[baseline, yshift=0.6ex]{\node[draw,circle,inner sep=1.1pt, font=\tiny] {4};} & \checkmark & \checkmark  & \textbf{28.47} & \textbf{32.68} & \textbf{0.8552} & \textbf{22.20}  \\

        % \bottomrule
        \bottomline
    \end{tabular}
    }
    % \vspace{-1mm}
    \caption{Analysis of FlexiTrack Instance: \(\textit{I}_{dn}\) represents a denoised instance generated from the Ground Truth (GT), whereas \(\textit{I}_{ft}\) refers to a FlexiTrack Instance.
    }
    % \vspace{-3mm}
    \label{tab: FlexiTrack_instance}
\end{table}

%% file: Table_AS_Comparison_of_DSSM_Block.tex
% \begin{table}[htbp]
\begin{table}[t]
    \centering
    \setlength{\tabcolsep}{8pt}     % 4pt
    \resizebox{\columnwidth}{!}{%
    \begin{tabular}{c|ccc|ccc}
        % \toprule
        \topline
        \rowcolor{mygray}  Exp. & $\mathcal{S}_5$ & $\mathcal{S}_4$ & $\mathcal{S}_3$ & HOTA$\uparrow$ & IDF1$\uparrow$ & OSPA$\downarrow$ \\
        % \midrule 
        \hline
        \tikz[baseline, yshift=0.6ex]{\node[draw,circle,inner sep=1.1pt, font=\tiny] {1};} & - &  -  &  -   & 27.61 & 31.21 & 0.8610  \\
        \tikz[baseline, yshift=0.6ex]{\node[draw,circle,inner sep=1.1pt, font=\tiny] {2};} & Conv &  Conv  &  Conv  & 24.93 & 27.88 & 0.8677  \\ 
        \tikz[baseline, yshift=0.6ex]{\node[draw,circle,inner sep=1.1pt, font=\tiny] {3};} & MLP &  MLP  &  MLP  & 23.19 & 26.51 & 0.8719  \\
        \tikz[baseline, yshift=0.6ex]{\node[draw,circle,inner sep=1.1pt, font=\tiny] {4};} & \checkmark &  \checkmark  &  \checkmark & 24.09 & 27.78 & 0.8755  \\ 
        \tikz[baseline, yshift=0.6ex]{\node[draw,circle,inner sep=1.1pt, font=\tiny] {5};}&  \checkmark  &    &    & 25.80 & 28.98 & 0.8660  \\
        \tikz[baseline, yshift=0.6ex]{\node[draw,circle,inner sep=1.1pt, font=\tiny] {6};}&    &    &  \checkmark  & 25.12 & 28.95 & 0.8636  \\

        % \midrule
        \hline
        \rowcolor{tabgray}\tikz[baseline, yshift=0.6ex]{\node[draw,circle,inner sep=1.1pt, font=\tiny] {7};} &  & \checkmark &    & \textbf{28.47} & \textbf{32.68} & \textbf{0.8552}  \\
        % \bottomrule
        \bottomline
    \end{tabular}
    }
    \vspace{-1mm}
    \caption{Ablation study on the DynamicSSM. $S_3$, $S_4$, and $S_5$ represent multi-scale features extracted from the backbone~\cite{He_2016_CVPR}. \emph{MLP} refers to multilayer perceptron layers, \emph{Conv} to convolutional layers. The symbol $\checkmark$ indicates the use of \emph{DynamicSSM} \ref{fig: CircularStatE Module}}

    % \vspace{-3mm}
    \label{tab:CSEM}
\end{table}

%% file: Table_AS_DynamicSSM.tex
% \begin{table}[htbp]
\begin{table}[t]
    \centering
    \setlength{\tabcolsep}{8pt}     % 4pt
    \resizebox{\columnwidth}{!}{%
    \begin{tabular}{c|ccc|ccc}
        % \toprule
        \topline
        \rowcolor{mygray}  Exp. & Dconv & SSM & Fusion & HOTA$\uparrow$ & IDF1$\uparrow$ & OSPA$\downarrow$  \\
        % \midrule
        \hline
         	 	 
        \tikz[baseline, yshift=0.6ex]{\node[draw,circle,inner sep=1.1pt, font=\tiny] {1};} & - & - & - & 27.61 & 31.21 & 0.8610\\ 	 	
        \tikz[baseline, yshift=0.6ex]{\node[draw,circle,inner sep=1.1pt, font=\tiny] {2};} & - & \checkmark & \checkmark &  26.31 & 29.70 & 0.8807   \\ 	 	
        \tikz[baseline, yshift=0.6ex]{\node[draw,circle,inner sep=1.1pt, font=\tiny] {3};} & \checkmark & - & \checkmark & 24.58 & 28.00 & 0.8777 \\ 
        \tikz[baseline, yshift=0.6ex]{\node[draw,circle,inner sep=1.1pt, font=\tiny] {4};} & \checkmark & \checkmark & - & 22.28 & 23.92 & 0.8878 \\ 

        % \midrule
        \hline
        \rowcolor{tabgray}\tikz[baseline, yshift=0.6ex]{\node[draw,circle,inner sep=1.1pt, font=\tiny] {5};} & \checkmark & \checkmark  & \checkmark & \textbf{28.47} & \textbf{32.68} & \textbf{0.8552}   \\
        % \bottomrule
        \bottomline
    \end{tabular}
    }
    %\vspace{-1mm}
\caption{Ablation study of the DynamicSSM Block, illustrating the contributions of each component to the overall performance: Dconv for deformable convolution (Eq.~\eqref{eq:dynamic_conv}), SSM for the state-space model (Eq.~\eqref{eq:ssm}), and Fusion for integrating residual features to enhance representation (Eq.~\eqref{eq:fusion}).}
    %\vspace{-3mm}
    \label{tab:dssm}
\end{table}

%% file: Table_AS_Epoch.tex
% \begin{table}[htbp]
\begin{table}[t]
    \centering
    \setlength{\tabcolsep}{8pt}     % 4pt
    \resizebox{\columnwidth}{!}{%
    \begin{tabular}{c|c|cccc}
        % \toprule
        \topline
        \rowcolor{mygray}  Exp. &  Epoch & HOTA$\uparrow$ & IDF1$\uparrow$ & OSPA$\downarrow$ & MOTA$\uparrow$   \\
        % \midrule
        \hline

        \tikz[baseline, yshift=0.6ex]{\node[draw,circle,inner sep=1.1pt, font=\tiny] {1};} & 5   &  24.20     & 28.36     & 0.8695    & 13.53     \\ 
        \tikz[baseline, yshift=0.6ex]{\node[draw,circle,inner sep=1.1pt, font=\tiny] {2};} & 6   &  28.47     & 32.68     & \textbf{0.8552}    & 22.20     \\
        \tikz[baseline, yshift=0.6ex]{\node[draw,circle,inner sep=1.1pt, font=\tiny] {3};} & 7   &   26.80     & 31.07     & 0.8687    & 19.29     \\ 
        \tikz[baseline, yshift=0.6ex]{\node[draw,circle,inner sep=1.1pt, font=\tiny] {4};} & 8   &  27.89     & 31.67     & 0.8735    & 21.82     \\
        \tikz[baseline, yshift=0.6ex]{\node[draw,circle,inner sep=1.1pt, font=\tiny] {5};} & 9   & 29.28     & 34.06     & 0.8672    & 27.25     \\ 
        \tikz[baseline, yshift=0.6ex]{\node[draw,circle,inner sep=1.1pt, font=\tiny] {6};} & 10   & 29.44     & 33.89     & 0.8663    & 23.05     \\
        \tikz[baseline, yshift=0.6ex]{\node[draw,circle,inner sep=1.1pt, font=\tiny] {7};} & 11   & 30.37     & 34.32     & 0.8573    & 22.79     \\ 
        \tikz[baseline, yshift=0.6ex]{\node[draw,circle,inner sep=1.1pt, font=\tiny] {8};} & 12   & 29.62     & 33.87     & 0.8604    & 25.01     \\ 
        \tikz[baseline, yshift=0.6ex]{\node[draw,circle,inner sep=1.1pt, font=\tiny] {9};} & 13   & 30.63     & 35.07     & 0.8759    & 29.63     \\
        \tikz[baseline, yshift=0.6ex]{\node[draw,circle,inner sep=0.5pt, font=\tiny] {10};} & 14   & 30.60     & 35.35     & 0.8737    & 29.69     \\ 
        \tikz[baseline, yshift=0.6ex]{\node[draw,circle,inner sep=0.5pt, font=\tiny] {11};} & 15   & 30.42     & 35.64     & 0.8781    & \textbf{30.72}     \\
        \tikz[baseline, yshift=0.6ex]{\node[draw,circle,inner sep=0.5pt, font=\tiny] {12};} & 16   & 30.58     & 35.08     & 0.8784    & 30.00     \\ 
        \tikz[baseline, yshift=0.6ex]{\node[draw,circle,inner sep=0.5pt, font=\tiny] {13};} & \textbf{17}   & \textbf{30.84}     & \textbf{35.66}     & 0.8792    & 30.09     \\
        \tikz[baseline, yshift=0.6ex]{\node[draw,circle,inner sep=0.5pt, font=\tiny] {14};} & 18   & 29.77     & 34.27     & 0.8893    & 28.81     \\ 
        \tikz[baseline, yshift=0.6ex]{\node[draw,circle,inner sep=0.5pt, font=\tiny] {15};} & 19   & 29.52     & 34.41     & 0.8948    & 29.51     \\
        \tikz[baseline, yshift=0.6ex]{\node[draw,circle,inner sep=0.5pt, font=\tiny] {16};} & 20   & 29.68     & 33.64     & 0.9017    & 29.21     \\

        \bottomline
    \end{tabular}
    }
    %\vspace{-1mm}
    \caption{Analysis of the impact of OmniTrack++$_{E2E}$ on tracking performance across different epochs.
    }
    %\vspace{-3mm}
    \label{tab:epoch}
\end{table}

%% file: Sec_Conclusion.tex
In this paper, we introduced \textbf{OmniTrack++}, a novel framework for panoramic multi-object tracking. Our approach unifies the End-to-End (E2E) and Tracking-by-Detection (TBD) paradigms via an adaptive Tracklet Management module. It reinforces temporal reasoning by leveraging a distortion-mitigating DynamicSSM block, trajectory-informed FlexiTrack Instances for stable short-term association, and a long-range ExpertTrack Memory to enhance identity preservation and recover fragmented trajectories.
To facilitate rigorous evaluation in this domain, we established the EmboTrack benchmark. 
By integrating the QuadTrack and BipTrack subsets, it provides a challenging testbed with heterogeneous robotic locomotion dynamics. The benchmark comprises $44$ panoramic sequences, over $26K$ annotated frames, and more than $600$ unique trajectories, setting a new standard for embodied panoramic MOT.
Extensive experiments validate our approach's effectiveness. On the challenging QuadTrack dataset, OmniTrack++ achieves state-of-the-art scores of $34.90$ HOTA in E2E mode and $36.08$ HOTA in TBD mode. 
Furthermore, on the JRDB benchmark, our E2E model reaches $25.50$ HOTA, significantly narrowing the performance gap to leading TBD methods and demonstrating the power of our feedback-driven design.
Despite these advances, challenges remain in scenarios with dense crowds and severe, prolonged occlusions. Future work will explore more explicit occlusion-aware modeling, enhance the underlying temporal reasoning within the closed-loop feedback system, and extend the framework toward long-term tracking in complex, real-world robotic environments.